\documentclass[a4paper,twoside]{article}

\usepackage{epsfig}
\usepackage{calc}
\usepackage{amssymb}
\usepackage{amstext}
\usepackage{amsmath}
\usepackage{amsthm}
\usepackage{multicol}
\usepackage{pslatex}
\usepackage{apalike}
\usepackage{algorithm2e}
\usepackage[bottom]{footmisc}
\usepackage{graphicx}
\usepackage{caption}
\usepackage{booktabs}
\usepackage{subfig}
\usepackage{mfirstuc}
\usepackage{longtable}
\usepackage{cite}
\usepackage{breakcites}
\usepackage{etoolbox}
\usepackage{changepage}
\usepackage{SCITEPRESS}     

\begin{document}
\makeatletter
\newcommand\notsotiny{\@setfontsize\notsotiny\@vipt\@viipt}
\patchcmd{\@citex}{,}{;}{}{}
\makeatother

\title{Fooling Neural Networks for Motion Forecasting via Adversarial Attacks}


\author{\authorname{Edgar Medina\sup{1}
		, Leyong Loh
		}
\affiliation{\sup{1}QualityMinds GmbH, Germany}
\email{{edgar.medina, leyong.loh}@qualityminds.de}
}

\keywords{Adversarial Attacks, Human Motion Prediction, 3D Deep Learning, Motion Analyses.}

\abstract{Human motion prediction is still an open problem, which is extremely important for autonomous driving and safety applications. Although there are great advances in this area, the widely studied topic of adversarial attacks has not been applied to multi-regression models such as GCNs and MLP-based architectures in human motion prediction. This work intends to reduce this gap using extensive quantitative and qualitative experiments in state-of-the-art architectures similar to the initial stages of adversarial attacks in image classification. The results suggest that models are susceptible to attacks even on low levels of perturbation. We also show experiments with 3D transformations that affect the model performance, in particular, we show that most models are sensitive to simple rotations and translations which do not alter joint distances. We conclude that similar to earlier CNN models, motion forecasting tasks are susceptible to small perturbations and simple 3D transformations.}

\onecolumn \maketitle \normalsize \setcounter{footnote}{0} \vfill

\section{\uppercase{Introduction}}
\label{sec:introduction}

Neural networks have demonstrated remarkable capabilities in achieving excellent performance in various 3D tasks, ranging from computer vision to robotics. Their capacity to process and analyze volumetric data, point clouds, or 3D meshes has been a driving force behind their success. Additionally, recent advancements in deep learning techniques, such as Convolutional Neural Networks (CNNs), Graph Convolutional Networks (GCNs), and Transformers, have substantially enhanced their ability to understand and manipulate 3D data. One notable aspect is their ability to generalize across different scales, orientations, and viewpoints has contributed to their robustness in handling diverse 3D tasks. This robustness is particularly valuable in scenarios where data may exhibit variations or perturbations, making neural networks a valuable tool for applications such as 3D object recognition, pose estimation, point cloud analysis, and others. In this work, we combine two different streams of artificial intelligence. On one side, deep neural networks have been applied to human motion prediction in different applications such as autonomous driving and bioinformatics \cite{Lyu2022a}, \cite{Wu2022a}. As mentioned in several papers, the surprising results of neural network architectures applied to motion prediction on 3D data structure surpassed classical approaches \cite{Lyu2022a}. Most of these methodologies are RNN-based family, GAN-based, mixed multi-linear nets, and GCNs. We focus mainly on the last two approaches limited to 3D pose forecasting on the Human 3.6M \cite{Ionescu2014} dataset, but could be easily extended to AMASS \cite{Mahmood2019} and 3DPW \cite{VonMarcard2018}. On the other hand, research in adversarial attacks has shown that neural networks are not robust enough for production due to how easily they are fooled with small perturbations and transformations \cite{Goodfellow2014}, \cite{Xiang2019}, \cite{Dong2017}. Although adversarial attacks have demonstrated success in fooling GCNs in recent years \cite{Chen2021, Liu2019a, Carlini2016a, Entezari2020, Tanaka2021, Diao2021}, these works focus on classification tasks, which are the predominant applications in machine learning. However, given the nature of our problem, many of these attack methods are not directly applicable to multi-output regression tasks. The investigation of adversarial attacks in human motion prediction is important. One example is in safety-critical autonomous driving systems, adversarial attacks can cause sensor failure in the pedestrian motion prediction module of the autonomous vehicle, consequently resulting in severe accidents.
This paper, to the best of our knowledge, presents the first effort of applying adversarial attacks in human motion prediction and intends to bridge this existing gap in the literature by conducting extensive experimentation on different state-of-the-art (SOTA) models.

It is important to highlight a limitation we encountered during the progression of our work is the absence of pre-trained models for several databases. Since the process of retraining these models using our own training codes may introduce a noteworthy degree of variability in our experiments, our work involves exclusively working with available pre-trained models and when corresponding source codes are accessible to train by ourselves. More concretely, the experiments involve applying gradient-based attacks and 3D spatial transformations to neural networks for multi-output regression tasks in human motion prediction. These experiments exclusively use white-box attacks because many methodologies adapted directly to multi-output regression do not yield effective results. In the preliminary stages of testing, white-box attacks serve as the initial choice due to their straightforward implementation. Conversely, black-box attacks introduce added intricacies when setting up experiments. Also, white-box attacks exhibit elevated success rates and facilitate the development of more refined and optimized attack tactics. White-box attacks demonstrate heightened success rates and enable the formulation of more precise and refined attack strategies, primarily due to their access to critical elements such as gradients, loss functions, and model architecture details. This stands in contrast to black-box attacks, which initially are resource-intensive strategies in order to obtain during the iterative process an understanding of the model. This potentially increases the complexity of designing effective optimization strategies and reduces the precision if this is not set appropriately. Finally, within the literature, there is an absence of references to black-box attacks employed specifically for multi-output regression tasks within graph models. We summarize our contributions as follows:

\begin{itemize}
    \item We perform extensive experiments on state-of-the-art models and evaluate the performance impact on the well-studied H3.6M dataset.
    \item We provide a methodology review for adversarial attacks applied to human motion prediction.
\end{itemize}

\section{\uppercase{Related Works}}

\subsection{Adversarial Attacks}
Two primary methodologies for adversarial attacks, namely White-box and Black-box methods, have been extensively studied in the literature. For the sake of simplicity and as explained in the introduction section, our experiments exclusively were set over white-box attacks and we excluded black-box methods in this analysis. In our literature review, we specifically focus on graph data, which exhibits parallels with the pose sequence data we employ. While we acknowledge the considerable impact of white-box attacks on performance, we also delve into the examination of 3D spatial transformations as adversarial attacks. These transformations will be subject to a detailed comparative analysis in subsequent sections.

\textbf{White-box:} One of the initial and enduring techniques in image perturbation is the Fast Gradient Signed Method (FGSM) \cite{Goodfellow2014}, which relies on computing the gradient through a single forward pass of the network. This method subsequently determines the gradient direction through a sign operation and scales it by an epsilon value before incorporating it into the current input. Furthermore, an iterative variant (I-FGSM) was introduced to mitigate pronounced effects when a large epsilon is used \cite{Goodfellow2014}. More recently, an iteration with a momentum factor (MI-FGSM) was introduced into the iterative algorithm, enhancing control over the gradient and reducing the perceptual visual impact of the attack \cite{Dong2017}. Iterative approaches have demonstrated their superiority over one-step methods in extensive experimentation as robust white-box algorithms, albeit at the expense of diminished transferability and increased computational demands \cite{Dong2019}. Later, Carlini and Wagner's method (C\&W) \cite{Carlini2016a} stands out as an unconstrained optimization-based technique known for its effectiveness, even when dealing with defense mechanisms. By harnessing first-order gradients, this algorithm seeks to minimize a balanced loss function that simultaneously minimizes the norm of the perturbation while maximizing the distance from the original input to evade detection. Another relevant method is DeepFool \cite{Moosavi-Dezfooli2015}, which focuses on identifying the smallest perturbation capable of causing misclassification by a neural network. DeepFool achieves this by linearizing a neural network and employing an iterative strategy to navigate the hyperplane in a simplified binary classification problem. The minimal perturbation required to alter the classifier's decision corresponds to the orthogonal projection of the input onto the hyperplane. DeepFool estimates the closest decision boundary for each class and iteratively adjusts the input in the direction of these boundaries \cite{Abdollahpourrostam2023}. Importantly, DeepFool is applicable to both binary and multi-class classification tasks.

While there is a wealth of research related to adversarial attacks in regression tasks \cite{gupta:hal-03527640}, \cite{Nguyen2018}, this area remains relatively unexplored, especially concerning multi-regression tasks or specialized domains such as 3D pose forecasting. Despite the significant progress in adversarial attacks, the majority of methods have mainly been applied to classification tasks. Therefore, we introduce a mathematical framework for regression tasks to address this gap in research.

\textbf{Adversarial attacks on graph data:} The output of pose estimation can be regarded as graph data, contingent on its structure and representation. Poses contain 2D or 3D positions, also called keypoints or joints, of the human body connected by a skeleton. Numerous prior studies within this domain have extensively explored gradients and their application in classification tasks \cite{Sun2018, Zhang2020, Fang2018, Zhu2019, Liu2019a, Carlini2016a, Entezari2020, Tanaka2021, Diao2021, Zugner2020}. In our work, we take these studies as inspiration to implement our approach in multi-output regression. In the next section, we provide more details about working with multiple real-valued outputs given that the target is not a binary output the methods must be adapted in this work. Furthermore, defense strategies previously studied in \cite{Chen2021} and \cite{Entezari2020a} showed that the attack effects can be reduced, however, these approaches are beyond the scope of this work.

\textbf{3D point cloud operations:} Taking inspiration from prior research on adversarial attacks in point clouds \cite{Dong2017}, we explore geometric transformations applied to 3D data points. Our objective is to manipulate these data points while preserving their distance distributions \cite{NEURIPS2021_82cadb06, Huang2022, Hamdi2019, Dong2017}. Notably, many of these previous works employ metrics such as Hausdorff distance and distribution metrics rather than MPJPE to quantify discrepancies between the adversarial input and the original input. Our experiments revolve around spatio-temporal affine transformations, including rotation, translation, and scaling, as a means to modify the pose sequences.

\subsection{3D Pose Forecasting}
In this section, we investigate various architectural paradigms, including GCNs, RNN-based models, and MLP-based models \cite{Lyu2022a}, that have been employed for human motion prediction. During the preprocessing stage, some models adopt Discrete Cosine Transforms (DCT) to transform the 3D input data into the frequency domain. This approach draws inspiration from prior work in graph processing, as discussed in \cite{Sun2018}. There are branches related to how the 3D input data is used: pre-processed and original.

Using a pre-processed input approach in the context of 3D pose prediction, DCT has been employed by \cite{Guo2022}, \cite{Ma2022}, \cite{Mao2020}, and \cite{Mao2021} to encode the input sequence. This encoding method is designed to capture the periodic body movements inherent in human motion. These studies showed substantial improvements in model predictions, surpassing the performance of previous RNN-based approaches. However, this operation is not only computationally expensive but also needs the use of Inverse DCT (IDCT) to revert the data back to the Euclidean space. Alternatively, other preprocessing approaches propose the replacement of input data or the aggregation of instantaneous displacements and displacement norm vectors \cite{Bouazizi2022b, Guo2022}. These methodologies have shown a notable reduction in error metrics. However, the pioneering network that, as far as our knowledge, directly incorporates the original 3D data into the model is called Space-Time-Separable Graph Convolutional Network (STS-GCN) \cite{Sofianos2021c}.

The current SOTA models in the field encompass MotionMixer \cite{Bouazizi2022b}, siMLPe \cite{Guo2022}, STS-GCN \cite{Sofianos2021c}, PGBIG \cite{Ma2022}, HRI \cite{Mao2020}, and MMA \cite{Mao2021}. MotionMixer and siMLPe are both MLP-based models that borrow the idea of Mixer architecture \cite{Tolstikhin2021} and apply it to the domain of human pose forecasting. STS-GCN and PGBIG are founded on GCNs. STS-GCN utilizes two successive GCNs to sequentially encode temporal and spatial pose data, while PGBIG introduces a multi-stage prediction framework with an iterative refining of the initial future pose estimate for improved prediction accuracy. HRI and MMA are GCN-based models augmented with attention modules. HRI introduces a motion attention mechanism based on DCT coefficients, which operate on sub-sequences of the input data. MMA takes a distinctive approach by employing an ensemble of HRI models at three different levels: full-body, body parts, and individual joints, to achieve enhanced prediction accuracy. In contrast to other SOTA models, MotionMixer adopts pose displacements as its input representation, whereas PGBIG, siMLPe, HRI, and MMA utilize DCT encoding for the input pose data. Additionally, recent works explored the hypothesis to include data augmentation in the training phase \cite{Maeda2021, Medina2024} or use rough form of interpretation in the adjacency matrices from the GCN layers \cite{Medina2024}. Tab. \ref{tab:MPJPE_per_class} and Tab. \ref{tab:AverageMPJPE} present the action-wise and average MPJPE results for the SOTA models on the Human3.6M dataset.

\section{\uppercase{Methodology}}

\begin{figure}
    \centering
    \includegraphics[width=\linewidth]{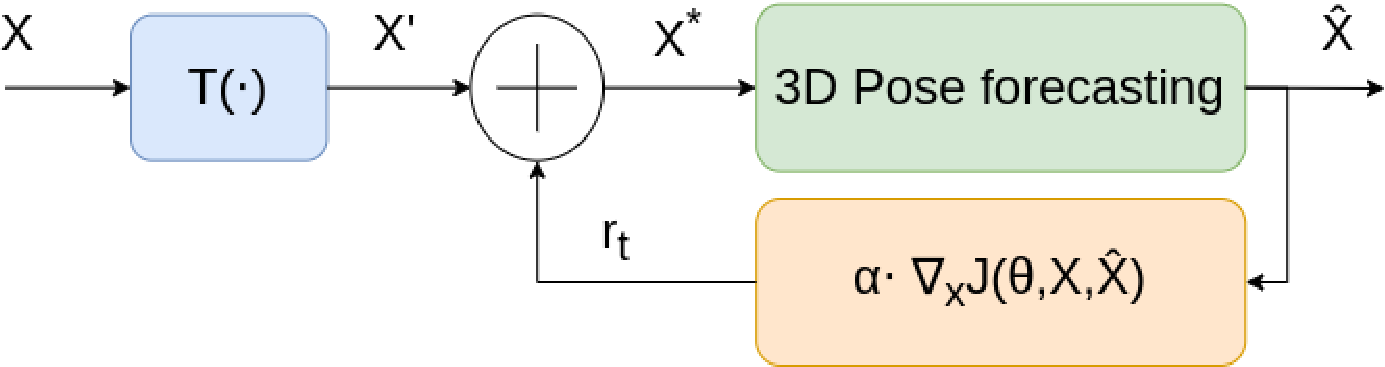}
    \caption{General diagram block for our experiments.}
    \label{fig:system}
\end{figure}

\subsection{Mathematical Notation}
Let $X$ be the input pose sequence and $\hat{X}$ the output pose sequence. Every pose sequence belongs to $\mathbb{R}^{T \times J \times D}$ where $T$ $J$, and $D$ are the temporal, joint domains and the 3D Euclidean positions respectively. Furthermore, we denote $X'$ and $X^*$ as the transformed and the adversarial examples for an input pose. This formulation process is visually depicted in Fig. \ref{fig:system}. It is important to observe that an identity transformation means that the transformed variable $X'$ would be identical to the original variable $X$.

\begin{table*}[]
\begin{adjustwidth}{-0.8cm}{}
\begin{center}
\small
\caption{Action-wise performance comparison of MPJPE for motion prediction on Human3.6M. We employ our evaluation pipeline only for the 1000ms models. A notable discrepancy exists between our evaluation results and those reported in the original papers. This variation is mainly due to the original papers using different strategies and frame evaluations.}
\begin{tabular}
{c|c@{\hspace{2pt}}c@{\hspace{2pt}}c@{\hspace{2pt}}c@{\hspace{2pt}}c@{\hspace{2pt}}c@{\hspace{2pt}}|c@{\hspace{2pt}}c@{\hspace{2pt}}c@{\hspace{2pt}}c@{\hspace{2pt}}c@{\hspace{2pt}}c@{\hspace{2pt}}|c@{\hspace{2pt}}c@{\hspace{2pt}}c@{\hspace{2pt}}c@{\hspace{2pt}}c@{\hspace{2pt}}c@{\hspace{2pt}}c@{\hspace{2pt}}c@{\hspace{2pt}}c@{\hspace{2pt}}c@{\hspace{2pt}}c@{\hspace{2pt}}c@{\hspace{2pt}}}

\toprule 
     & 
    \multicolumn{6}{c}{Walking} & 
    \multicolumn{6}{|c}{Eating} &
    \multicolumn{6}{|c}{Smoking} \\

    Time (ms) &
    80 & 160 & 320 & 400 & 560 & 1000 &
    80 & 160 & 320 & 400 & 560 & 1000 &
    80 & 160 & 320 & 400 & 560 & 1000 \\
    \midrule

    STS-GCN \cite{Sofianos2021c} & 18.0 & 32.9 & 46.7 & 53.4 & 58.0 & 70.2 & 12.1 & 23.3 & 36.8 & 44.3 & 57.4 & 82.6 & 13.0 & 16.4 & 37.2 & 44.6 & 55.5 & 76.1 \\

    PGBIG \cite{Ma2022} & 14.5 & 34.5 & 66.7 & 79.0 & 96.7 & 111.2 & 7.93 & 19.2 & 38.9 & 48.0 & 63.8 & 88.7 & 8.55 & 20.0 & 39.7 & 48.1 & 62.0 & 85.5 \\
    
    MotionMixer \cite{Bouazizi2022b} & 14.4 & 27.0 & 46.2 & 52.7 & 58.7 & 66.1 & 8.5 & 17.3 & 33.5 & 41.3 & 54.4 & 79.9 & 9.0 & 17.9 & 34.3 & 41.7 & 53.2 & 74.3 \\
    
    siMLPe \cite{Guo2022} & 12.0 & 20.5 & 34.8 & 40.2 & 48.7 & 57.3 & 9.9 & 16.2 & 30.4 & 37.6 & 53.4 & 79.1 & 10.5 & 17.0 & 31.5 & 38.0 & 51.4 & 73.9 \\

    HRI \cite{Mao2020} & 10.0 & 19.5 & 34.2 & 39.8 & 47.4 & 58.1 & 6.4 & 14.0 & 28.7 & 36.2 & 50.0 & 75.6 & 7.0 & 14.9 & 29.9 & 36.4 & 47.6 & 69.5 \\
 
    MMA \cite{Mao2021} & \textbf{9.9} & \textbf{19.3} & \textbf{33.8} & \textbf{39.1} & \textbf{46.2} & \textbf{57.2} & \textbf{6.2} & \textbf{13.7} & \textbf{28.1} & \textbf{35.3} & \textbf{48.7} & \textbf{73.8} & \textbf{6.8} & \textbf{14.5} & \textbf{29.0} & \textbf{35.5} & \textbf{46.5} & \textbf{68.8} \\

    \midrule 
     &
    
    \multicolumn{6}{c}{Discussion} &
    \multicolumn{6}{|c}{Directions} & 
    \multicolumn{6}{|c}{Greeting} \\

    Time (ms) &
    80 & 160 & 320 & 400 & 560 & 1000 &
    80 & 160 & 320 & 400 & 560 & 1000 &
    80 & 160 & 320 & 400 & 560 & 1000 \\
    \midrule 

    STS-GCN \cite{Sofianos2021c} & 17.1 & 33.2 & 58.6 & 71.3 & 91.1 & 118.9 & 13.9 & 29.3 & 52.8 & 64.0 & 79.9 & 109.6 & 20.8 & 40.7 & 72.2 & 85.9 & 106.3 & \textbf{136.1} \\

    PGBIG \cite{Ma2022} & 12.1 & 28.7 & 58.5 & 71.6 & 93.6 & 123.9 & 9.1 & 23.0 & 49.8 & 61.4 & 77.7 & 108.9 & 16.2 & 38.0 & 74.6 & 88.6 & 111.1 & 143.8 \\
    
    MotionMixer \cite{Bouazizi2022b} & 12.7 & 27.1 & 56.8 & 70.2 & 91.7 & 123.8 & 9.0 & 20.9 & 48.1 & 60.2 & 76.9 & 110.1 & 16.6 & 35.5 & 72.7 & 87.5 & 110.8 & 145.6 \\

    siMLPe \cite{Guo2022} & 12.5 & 24.6 & 52.2 & 65.2 & 87.8 & 118.8 & 11.2 & 20.4 & 45.7 & 57.2 & 76.3 & 110.1 & 15.1 & 30.2 & 63.5 & 77.8 & 101.3 & 139.3 \\

    HRI \cite{Mao2020} & 10.2 & 23.4 & 52.1 & 65.4 & 86.6 & 119.8 & 7.4 & 18.4 & 44.5 & 56.5 & 73.9 & 106.5 & 13.7 & 30.1 & 63.8 & 78.1 & 101.9 & 138.8 \\
 
    MMA \cite{Mao2021} & \textbf{9.9} & \textbf{22.9} & \textbf{51.0} & \textbf{64.0} & \textbf{85.3} & \textbf{117.8} & \textbf{7.2} & \textbf{18.0} & \textbf{43.3} & \textbf{55.0} & \textbf{72.3} & \textbf{105.8} & \textbf{13.6} & \textbf{30.0} & \textbf{63.2} & \textbf{77.5} & \textbf{101.0} & 137.9 \\

    \midrule 
     &
    
    \multicolumn{6}{c}{Phoning}&
    \multicolumn{6}{|c}{Posing} &
    \multicolumn{6}{|c}{Purchases} \\

    Time (ms) &
    80 & 160 & 320 & 400 & 560 & 1000 &
    80 & 160 & 320 & 400 & 560 & 1000 &
    80 & 160 & 320 & 400 & 560 & 1000 \\
    \midrule 

    STS-GCN \cite{Sofianos2021c} & 14.5 & 27.3 & 45.7 & 55.4 & 73.1 & 108.3 & 19.0 & 38.9 & 71.7 & 89.2 & 119.7 & 178.4 & 20.9 & 41.4 & 71.8 & 86.0 & 106.8 & 141.0\\

    PGBIG \cite{Ma2022} & 10.2 & 23.5 & 48.5 & 59.3 & 77.6 & 114.4 & 12.3 & 30.6 & 66.2 & 82.1 & 111.0 & 173.8 & 14.8 & 34.2 & 65.5 & 78.4 & 99.3 & 136.7\\
    
    MotionMixer \cite{Bouazizi2022b} & 10.5 & 21.2 & 43.6 & 54.2 & 72.6 & 110.1 & 12.6 & 28.3 & 65.0 & 82.5 & 113.4 & 181.3 & 15.3 & 33.5 & 67.8 & 81.7 & 102.6 & 143.7\\
    
    siMLPe \cite{Guo2022} & 11.8 & 19.9 & 39.8 & 49.4 & 68.2 & \textbf{103.7} & 13.4 & 25.8 & 58.8 & 74.8 & 105.9 & \textbf{170.0} & 15.6 & 29.9 & 60.0 & 73.5 & 96.5 & 135.4\\

    HRI \cite{Mao2020} & 8.6 & 18.3 & 39.0 & 49.2 & 67.4 & 105.0 & 10.2 & 24.2 & 58.5 & 75.8 & 107.6 & 178.2 & 13.0 & 29.2 & 60.4 & 73.9 & 95.6 & 134.2 \\
 
    MMA \cite{Mao2021} & \textbf{8.5} & \textbf{18.0} & \textbf{38.3} & \textbf{48.4} & \textbf{66.6} & 104.1 & \textbf{9.8} & \textbf{23.7} & \textbf{58.0} & \textbf{75.1} & \textbf{105.8} & 171.5 & \textbf{12.8} & \textbf{28.7} & \textbf{59.5} & \textbf{72.8} & \textbf{94.6} & \textbf{133.6} \\

    \midrule 
     & 

    \multicolumn{6}{c}{Sitting} &
    \multicolumn{6}{|c}{Sitting Down} &
    \multicolumn{6}{|c}{Taking Photo} \\

    Time (ms) &
    80 & 160 & 320 & 400 & 560 & 1000 &
    80 & 160 & 320 & 400 & 560 & 1000 &
    80 & 160 & 320 & 400 & 560 & 1000 \\
    \midrule 

    STS-GCN \cite{Sofianos2021c} & 16.0 & 30.1 & 51.9 & 63.9 & 84.7 & 121.4 & 23.9 & 42.9 & 68.9 & 81.5 & 105.2 & 148.4 & 16.2 & 31.3 & 52.1 & 63.4 & 84.2 & 126.3\\

    PGBIG \cite{Ma2022} & 10.0 & 22.2 & 45.5 & 56.3 & 76.2 & \textbf{114.4} & 15.8 & 33.5 & 61.8 & 74.1 & 98.5 & 143.3 & 9.3 & 21.3 & 44.6 & 55.5 & 75.7 & 116.1 \\
    
    MotionMixer \cite{Bouazizi2022b} & 10.8 & 22.4 & 46.5 & 57.8 & 78.0 & 116.4 & 17.1 & 34.6 & 64.3 & 77.7 & 103.3 & 149.6 & 9.6 & 20.3 & 43.5 & 54.5 & 75.3 & 118.3 \\
    
    siMLPe \cite{Guo2022} & 13.2 & 22.2 & 45.5 & 56.8 & 79.3 & 118.2 & 18.3 & 32.4 & 59.8 & 72.3 & 99.2 & 144.8 & 12.5 & 20.5 & 41.6 & 52.1 & 73.8 & \textbf{114.1} \\

    HRI \cite{Mao2020} & 9.3 & 20.1 & 44.3 & 56.0 & 76.4 & 115.9 & 14.9 & 30.7 & 59.1 & 72.0 & 97.0 & 143.6 & 8.3 & 18.4 & 40.7 & 51.5 & 72.1 & 115.9 \\
 
    MMA \cite{Mao2021} & \textbf{9.1} & \textbf{19.7} & \textbf{43.7} & \textbf{55.5} & \textbf{75.8} & 114.6 & \textbf{14.7} & \textbf{30.4} & \textbf{58.5} & \textbf{71.4} & \textbf{96.2} & \textbf{142.0} & \textbf{8.2} & \textbf{18.1} & \textbf{40.2} & \textbf{51.1} & \textbf{71.8} & 114.6 \\
    
    \midrule 
     & 
    \multicolumn{6}{c}{Waiting} & 
    \multicolumn{6}{|c}{Walking Dog} &
    \multicolumn{6}{|c}{Walking Together} \\

    Time (ms) &
    80 & 160 & 320 & 400 & 560 & 1000 &
    80 & 160 & 320 & 400 & 560 & 1000 &
    80 & 160 & 320 & 400 & 560 & 1000 \\
    \midrule 
    
    STS-GCN \cite{Sofianos2021c} & 15.9 & 31.5 & 52.3 & 63.4 & 80.8 & 113.6 & 29.2 & 53.3 & 84.2 & 96.1 & 115.4 & 151.5 & 15.5 & 28.2 & 42.3 & 49.9 & 58.9 & 72.5 \\

    PGBIG \cite{Ma2022} & 10.6 & 25.2 & 53.1 & 65.2 & 83.8 & 113.1 & 22.5 & 47.2 & 81.1 & 93.6 & 113.8 & 151.4 & 11.5 & 28.0 & 55.2 & 66.7 & 83.0 & 99.0 \\
    
    MotionMixer \cite{Bouazizi2022b} & 10.8 & 22.8 & 49.1 & 61.0 & 80.3 & 113.4 & 24.2 & 47.2 & 81.0 & 93.0 & 111.6 & 153.8 & 11.6 & 23.1 & 43.5 & 51.7 & 61.2 & 69.9 \\
    
    siMLPe \cite{Guo2022} & 11.5 & 20.6 & 43.7 & 54.2 & 74.0 & 106.1 & 22.3 & 40.7 & 72.6 & 85.0 & 108.7 & 145.0 & 11.8 & 19.9 & 35.7 & 42.1 & 53.7 & 64.6 \\

    HRI \cite{Mao2020} & 8.7 & 19.2 & 43.4 & 54.9 & 74.5 & 108.2 & 20.1 & 40.3 & 73.3 & 86.3 & 108.2 & 146.8 & 8.9 & 18.4 & 35.1 & 41.9 & 52.7 & 64.9 \\
 
    MMA \cite{Mao2021} & \textbf{8.4} & \textbf{18.8} & \textbf{42.5} & \textbf{53.8} & \textbf{72.6} & \textbf{104.8} & \textbf{19.6} & \textbf{39.5} & \textbf{71.8} & \textbf{84.3} & \textbf{105.1} & \textbf{142.1} & \textbf{8.5} & \textbf{17.9} & \textbf{34.4} & \textbf{41.2} & \textbf{51.3} & \textbf{63.3} \\
    \bottomrule
\end{tabular}
\label{tab:MPJPE_per_class}
\end{center}
\end{adjustwidth}
\end{table*}

\begin{table}[h]
\begin{adjustwidth}{-1cm}{}
    \small
    \caption{Average MPJPE for H3.6M dataset}
    \centering
    \begin{tabular}{c|c@{\hspace{1.5pt}}c@{\hspace{1.5pt}}c@{\hspace{1.5pt}}c@{\hspace{1.5pt}}c@{\hspace{1.5pt}}c@{\hspace{1.5pt}}}
    \toprule
    &
    \multicolumn{6}{c}{Average} \\
    Time (ms) & 80 & 160 & 320 & 400 & 560 & 1000 \\
    \midrule 
     STS-GCN \cite{Sofianos2021c} & 17.7 & 33.9 & 56.3 & 67.5 & 85.1 & 117.0\\
     PGBIG \cite{Ma2022} & 12.4 & 28.6 & 56.7 & 68.5 & 88.2 & 121.6\\ 
     MotionMixer \cite{Bouazizi2022b} & 12.8 & 26.6 & 53.1 & 64.5 & 82.9 & 117.1\\
     siMLPe \cite{Guo2022} & 13.4 & 24.0 & 47.7 & 58.4 & 78.6 & 112.0 \\
     HRI \cite{Mao2020} & 10.4 & 22.6 & 47.1 & 58.3 & 77.3 & 112.1 \\
     MMA \cite{Mao2021} & \textbf{10.2} & \textbf{22.2} & \textbf{46.4} & \textbf{57.3} & \textbf{76.0} & \textbf{110.1} \\
    \end{tabular}
    \label{tab:AverageMPJPE}
\end{adjustwidth}
\end{table}

\subsection{\capitalisewords{Gradient-based Methods}}
We first detail a commonly used non-targeted attack called FGSM algorithm. FGSM finds an adversarial example $X^*$ that maximizes the loss function $L(\theta,x,y)$ composed by $L_\infty$ norm of the difference which is limited to a small $\epsilon$. This is expressed in Eq. \ref{eq:fgsm}.

\begin{equation}\label{eq:fgsm}
    X^* = X + \epsilon \cdot sign(\nabla_x L(\theta,x,y)) , \quad \lVert X^* - X \rVert _ \infty \leq \epsilon
\end{equation}

The iterative extension of FGSM, known as IFGSM, applies the fast gradient method for multiple iterations, introducing incremental changes controlled by a scaling factor denoted as $\alpha$. Consequently, the adversarial example $X^*$ is updated iteratively for a total of $t$ times as described in  Eq. \ref{eq:ifgsm}, where $X^t$ represents the adversarial example at the t-th iteration. It is essential to emphasize that, as outlined in \cite{Dong2017}, $X^*$ must satisfy the boundary condition imposed by $\epsilon$ to ensure minimal perturbation, as denoted by $\lVert X^* - X \rVert \leq \epsilon$ within the algorithm. The choice of norm for this constraint typically includes the $L_0$, $L_2$, and $L\infty$ norms. An alternative representation for the perturbation factor $\alpha$ can be derived by setting $\alpha = \epsilon / T$, where $T$ is the total number of iterations.

\begin{equation}\label{eq:ifgsm}
    X^*_{0} = X, \quad  X^*_{t+1} = X^*_t + \alpha \cdot sign(\nabla_{x^*_t} L(\theta,x,y))
\end{equation}

A more sophisticated variant of IFGSM has been proposed to improve the transferability of adversarial examples by incorporating momentum within the iterative process. The update procedure is shown in Eqs. \ref{eq:mifgsm1} and \ref{eq:mifgsm2}, where $g_t$ gathers the gradient information in the t-th iteration, subject to a decay factor $\mu$.

\begin{equation}\label{eq:mifgsm1}
    g_{t+1} = \mu \cdot g_t + \frac{\nabla_{x^*_t} L(\theta,x,y)}{\lVert \nabla_{x^*_t} L(\theta,x,y) \rVert _ 1}
\end{equation}
\begin{equation}\label{eq:mifgsm2}
    X^*_{t+1} = X^*_t + \alpha \cdot sign(g_{t+1})
\end{equation}

The DeepFool approach merges a linearization strategy with the vector projection of a given sample. This projection corresponds to the minimum distance required to cross the hyperplane in the context of binary classification. Subsequently, this approach has been extended to multi-class classification scenarios. The noise introduced is described as a vector projection originating from a point $x_0$, oriented in the direction of the hyperplane's normal vector $W$, with the distance serving as its magnitude. This mathematical representation is expressed in Eq. \ref{eq:deepfool}. However, with the iterative application of noise, its representation evolves as shown in Eq. \ref{eq:deepfool1}.

\begin{equation}\label{eq:deepfool}
    r_*(x_0) = - \frac{f(x_0)}{\lVert w \rVert_2^2} w
\end{equation}

\begin{equation}\label{eq:deepfool1}
    r_i \leftarrow - \frac{f(x_i)}{\lVert \nabla f(x_i) \rVert_2^2}\nabla f(x_i)
\end{equation}

\begin{figure}
    \centering
    \includegraphics[width=0.75\linewidth]{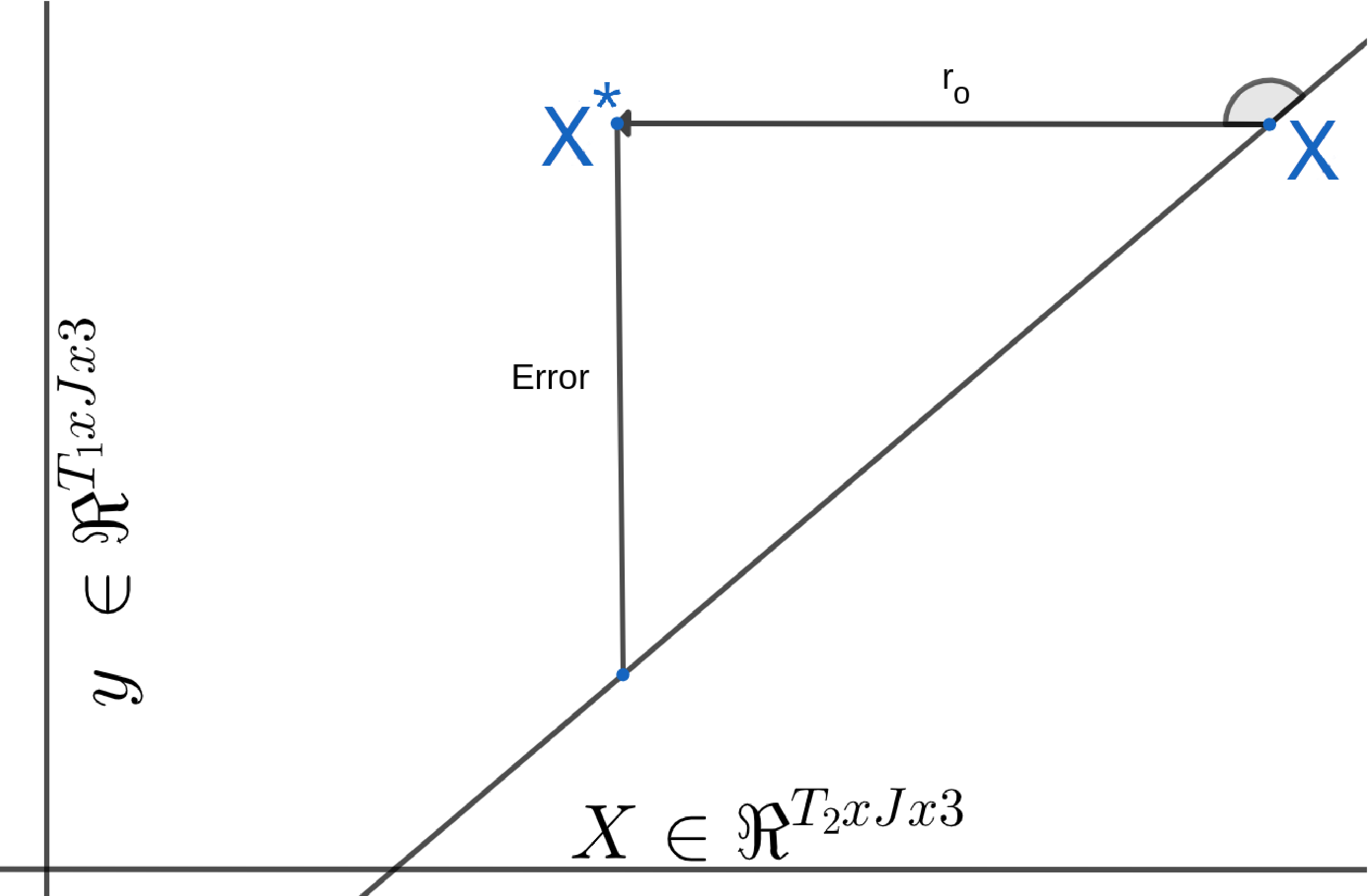}
    \caption{Visual interpretation of an algorithm for regression tasks.}
    \label{fig:deepfool_reg}
\end{figure}

Our methodology comprises two primary steps. First, we establish a neural network linearization approach, similar to previous methodologies explored in works such as \cite{Moosavi-Dezfooli2015, gupta:hal-03527640, Nguyen2018}. Second, we introduce a noise vector to the point $x_0$, strategically positioned to be in very close proximity to the hyperplane $\Pi :W^TX$. This is illustrated in Fig. \ref{fig:deepfool_reg}. This noise vector must have a horizontal orientation to ensure that an input changes, transitioning from $X \in \mathbb{R}^{T_2xJx3}$ to $\hat{X} \in \mathbb{R}^{T_2xJx3}$, does not induce any alterations in the output $y \in \mathbb{R}^{T_1xJx3}$. This condition can be mathematically expressed as a straightforward regression problem, namely $y=f_W(x_0)$, where typically $T_1$ and $T_2$ differ. Also, it becomes evident from the figure that the error denoted as $E$ is directly proportional to the magnitude of the noise vector $\lVert r_0 \rVert$. For instance, when considering an angle of 45° within a right triangle, the error magnitude aligns with the magnitude of the noise vector. This observation means that networks with heightened susceptibility, particularly from a numerical instability perspective, are more likely to exhibit larger errors.

\subsection{\capitalisewords{Spatio-Temporal Transformations}}
In contrast to adding noise into the pose sequences, we propose an alternative approach involving the application of 3D transformations, specifically affine transformations, to evaluate the models. The general form of an affine transformation in homogeneous coordinates is mathematically represented in Eq. \ref{eq:transformation}, where $A$ represents the affine matrix, and $t$ is the translation vector. The matrix $A$ implicitly contains both the rotation matrix and the scaling factor through a matrix multiplication operation between these two matrices. The output of the transformation, denoted as $X'$, is obtained after operating the transformation $T(\cdot)$. Hence, the final adversarial example is derived through the addition of noise $r_t$, as illustrated in Fig. \ref{fig:system}. This process is formally expressed in Eqs. \ref{eq:overall_transformation_1} and \ref{eq:overall_transformation_2}. It is worth noting that the transformation operation $T(\cdot)$ may behave as an identity matrix in certain cases.

\begin{equation}\label{eq:transformation}
    H = \begin{bmatrix}
    A & t \\
    0^T & 1
    \end{bmatrix} \quad , where \quad \mathrm{det}A \neq 0
\end{equation}

\begin{equation}\label{eq:overall_transformation_1}
    X' = T(X)
\end{equation}

\begin{equation}\label{eq:overall_transformation_2}
    X^* = X' + r_t
\end{equation}

\section{\uppercase{Experimentation}}

\subsection{\capitalisewords{Datasets}}
We conducted experiments on the Human 3.6M \cite{Ionescu2014} dataset. This dataset consists of a diverse set of 15 distinct actions performed by 7 different actors. To facilitate consistent processing, we downsampled the frame rate to 25Hz and employed a 22-joint configuration same as in \cite{Mao2020} and \cite{Fu2022a}. It means subjects 1, 6, 7, 8, and 9 were set for training, subject 11 for validation, and subject 5 for testing.

\subsection{Results}
We have opted to select SOTA models for our experiments. To ensure a fair comparison, we have uniformly implemented all these models within our evaluation pipeline and have evaluated them using a consistent standard metric. These experiments serve to analyze the impact of adversarial attacks and spatial transformations on the performance of these models, thereby providing valuable insights into the strengths and limitations of various architectures. We also attempted to conduct experiments with AMASS and 3DPW datasets. However, we encountered a challenge as there were no training codes or pre-trained models available for all the models. Also, conducting a fair and meaningful comparison across all models under these conditions was rendered infeasible. Therefore, only STS-GCN, MotionMixer and siMLPe are evaluated on AMASS and 3DPW.

\textbf{Metric}. In our evaluations, we employ the Mean Per Joint Position Error (MPJPE) metric, which is a widely accepted evaluation measure commonly used in previous works \cite{Lyu2022a, Dang2021, Liu2021b} to compare two pose sequences. The MPJPE metric is defined in Eq. \eqref{mpjpe}.

\begin{equation}\label{mpjpe}
    \mathcal{L}_{\text{MPJPE}} = \frac{1}{J \times T} \sum^{T}_{t=1}\sum^{J}_{j=1}{ \lVert \hat{x}_{j,t} - x_{j,t} \rVert _2}
\end{equation}

\begin{figure}
    \centering
\includegraphics[trim=10 0 10 25, clip, width=\linewidth]{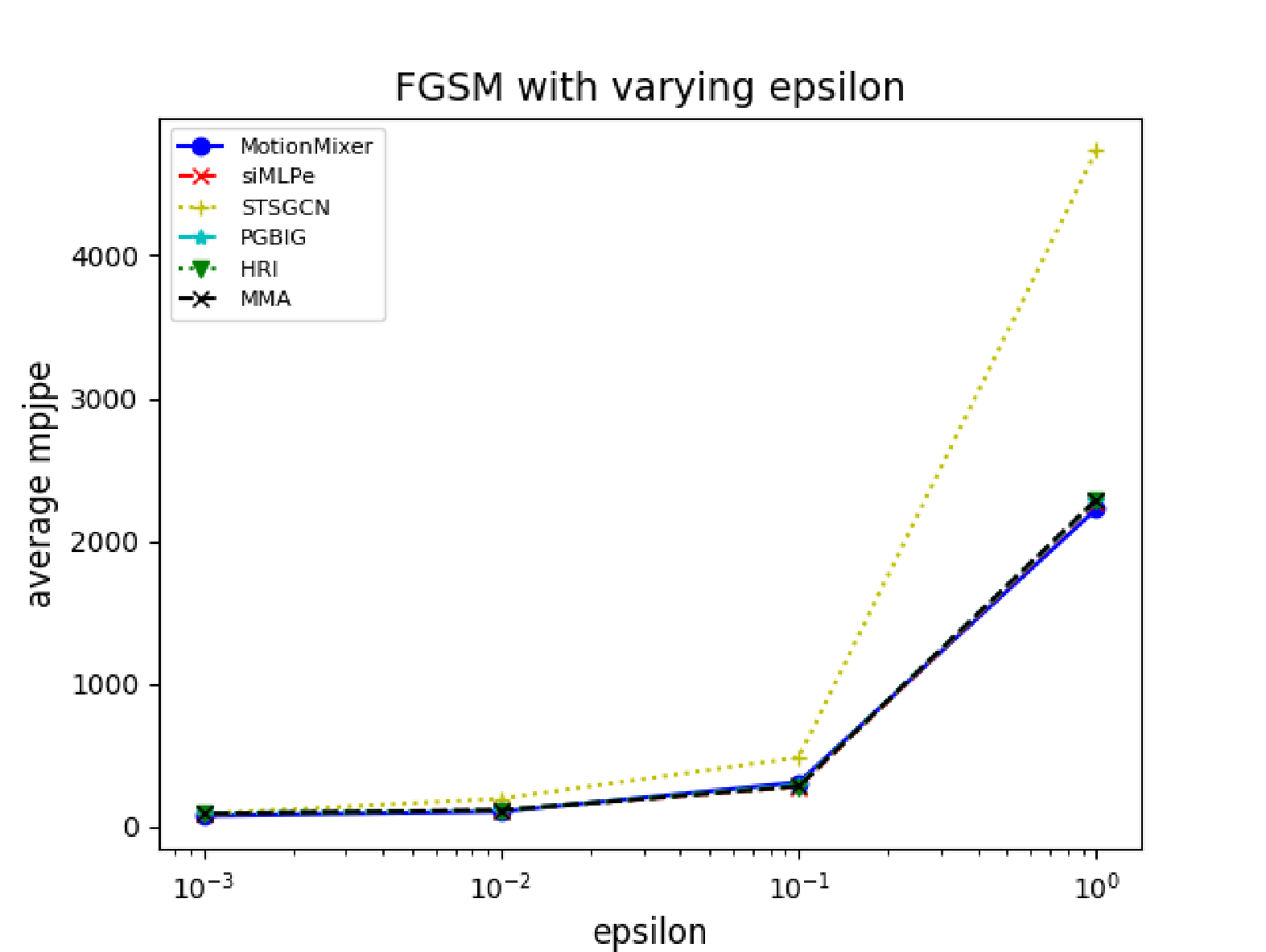}
    \caption{Result of FGSM with increasing epsilon on average MPJPE.}
    \label{fig:FGSM_epsilon}
\end{figure}

\begin{table*}[ht]
\begin{adjustwidth}{0.0cm}{}
\small
    \caption{Comparison of adversarial attacks on average MPJPE for H3.6M. The arrows denote superior results. ($\epsilon = 0.001$)}
    \centering
    \begin{tabular}{cc@{\hspace{5pt}}c@{\hspace{5pt}}c@{\hspace{5pt}}c@{\hspace{5pt}}c@{\hspace{5pt}}c@{\hspace{5pt}}c@{\hspace{5pt}}}
    \toprule 
    Model & IFGSM ($\uparrow$) & $\Delta s$ ($\downarrow$) & MIFGSM ($\uparrow$) & $\Delta s$ ($\downarrow$) & DeepFool ($\uparrow$) & $\Delta s$ ($\downarrow$) & w/o \\
    \midrule 
     STS-GCN & 86.1 (+16\%) & 1.3 & 86.1 (+16\%) & 1.3 & 89.5 (+20\%) & 2.6 & 74.7 \\
     PGBIG & 96.9 (+28\%) & 0.3 & 96.2 (+27\%) & 0.3 & 86.7 (+15\%) & 0.9 & 75.2 \\ 
     MotionMixer & \textbf{72.3 (+1\%)} & \textbf{0.02} & \textbf{72.3 (+1\%)} & \textbf{0.02} & \textbf{101.9 (+43\%)} & \textbf{2.6} & 71.2 \\
     siMLPe & 131.3 (+94\%) & 1.1 & 133.1 (+97\%) & 1.2 & 86.9 (+29\%) & 0.9 & 67.5 \\ 
     HRI & 115.2 (+73\%) & 1.2 & 113.8 (+71\%) & 1.2 & 82.1 (+24\%) & 0.7 & 66.4 \\ 
     MMA & 107.1 (+64\%) & 1.2 & 105.9 (+62\%) & 1.2 & 79.7 (+22\%) & 0.7 & 65.3 \\
    \bottomrule
    \end{tabular}
    \label{tab:Attacks_H36M_epsilon_0.001}
\end{adjustwidth}
\end{table*}

\begin{table*}[ht]
\begin{adjustwidth}{0.0cm}{}
\small
    \caption{Comparison of adversarial attacks on average MPJPE for H3.6M. The arrows denote superior results. ($\epsilon = 0.01$)}
    \centering
    \begin{tabular}{cc@{\hspace{5pt}}c@{\hspace{5pt}}c@{\hspace{5pt}}c@{\hspace{5pt}}c@{\hspace{5pt}}c@{\hspace{5pt}}c@{\hspace{5pt}}}
    \toprule 
    Model & IFGSM ($\uparrow$) & $\Delta s$ ($\downarrow$) & MIFGSM ($\uparrow$) & $\Delta s$ ($\downarrow$) & DeepFool ($\uparrow$) & $\Delta s$ ($\downarrow$) & w/o \\
    \midrule 
     STS-GCN & 239.5 (+220\%) & 13.1 & 236.2 (+216\%) & 13.5 & 89.5 (+20\%) & 2.6 & 74.7\\
     PGBIG & 173.7 (+131\%) & 2.3 & 178.2 (+137\%) & 2.6 & 86.7 (+15\%) & 0.9 & 75.2 \\ 
     MotionMixer & \textbf{84.1 (+18\%)} & \textbf{0.2} &  \textbf{83.9 (+17\%)} & \textbf{0.2} & \textbf{101.9 (+43\%)} & \textbf{2.6} & 71.2 \\
     siMLPe & 281.1 (+316\%) & 10.7 & 296.6 (+339\%) & 11.5  & 86.9 (+29\%) & 0.9 & 67.5 \\ 
     HRI & 264.1 (+297\%) & 11.5 & 274.0 (+312\%) & 12.1 & 82.1 (+24\%) & 0.7 & 66.4 \\ 
     MMA & 250.8 (+284\%) & 11.7 & 259.7 (+297\%) & 12.2 & 79.7 (+22\%) & 0.7 & 65.3 \\
    \bottomrule
    \end{tabular}
    \label{tab:Attacks_H36M_epsilon_0.01}
\end{adjustwidth}
\end{table*}

\begin{table*}[h!]
\begin{adjustwidth}{0.0cm}{}
\small
    \caption{Comparison of adversarial attacks on the average MPJPE for AMASS. The arrows denote superior results. ($^\ast$) model was trained by us. ($\epsilon = 0.01$)}
    \centering
    \begin{tabular}{cc@{\hspace{5pt}}c@{\hspace{5pt}}c@{\hspace{5pt}}c@{\hspace{5pt}}c@{\hspace{5pt}}c@{\hspace{5pt}}c@{\hspace{5pt}}}
    \toprule 
    Model & IFGSM ($\uparrow$) & $\Delta s$ ($\downarrow$) & MIFGSM ($\uparrow$) & $\Delta s$ ($\downarrow$) & DeepFool ($\uparrow$) & $\Delta s$ ($\downarrow$) & w/o \\
    \midrule 
     STS-GCN & 194.8 (+78\%) & 12.2 & 194.8 (+78\%) & 12.2 & 119.6 (+9\%) & 4.3 & 109.6 \\
     MotionMixer & \textbf{89.1 (+1\%)} & \textbf{0.17} & \textbf{89.1 (+1\%)} & \textbf{0.17} & \textbf{89.0 (+1\%)} & \textbf{2.6} & 88.5  \\
     siMLPe $^\ast$ & 189.8 (+406\%) & 12.5 & 190.2 (+407\%) & 12.5 & 66.9 (+78\%) & 1.3 & 37.5 \\ 
    \bottomrule
    \end{tabular}
    \label{tab:Attacks_AMASS_epsilon_0.01}
\end{adjustwidth}
\end{table*}

\begin{table*}[h!]
\begin{adjustwidth}{0.0cm}{}
\small
    \caption{Comparison of adversarial attacks on average MPJPE for 3DPW. The arrows denote superior results. ($^\ast$) model was trained by us. ($\epsilon = 0.01$)}
    \centering
    \begin{tabular}{cc@{\hspace{5pt}}c@{\hspace{5pt}}c@{\hspace{5pt}}c@{\hspace{5pt}}c@{\hspace{5pt}}c@{\hspace{5pt}}c@{\hspace{5pt}}}
    \toprule 
    Model & IFGSM ($\uparrow$) & $\Delta s$ ($\downarrow$) & MIFGSM ($\uparrow$) & $\Delta s$ ($\downarrow$) & DeepFool ($\uparrow$) & $\Delta s$ ($\downarrow$) & w/o \\
    \midrule 
     STS-GCN & 190.4 (+79\%) & 13.7 & 190.2 (+79\%) & 13.7 & 115.5 (+9\%) & 4.3 & 106.1 \\
     MotionMixer & \textbf{69.5 (+1\%)} & \textbf{0.19} & \textbf{69.5 (+1\%)} & \textbf{0.19} & \textbf{69.0 (+0\%)} & \textbf{2.5} & 69.0 \\
     siMLPe $^\ast$  & 189.5 (+358\%) & 12.4 & 190.0 (+360\%) & 12.3 & 63.7 (+54\%) & 1.3 & 41.3 \\ 
    \bottomrule
    \end{tabular}
    \label{tab:Attacks_3DPW_epsilon_0.01}
\end{adjustwidth}
\end{table*}

\textbf{Quantitative results}.
We conducted adversarial attacks using the IFGSM, MIFGSM, and DeepFool techniques on the SOTA models. Tabs. \ref{tab:Attacks_H36M_epsilon_0.001} and \ref{tab:Attacks_H36M_epsilon_0.01} present the outcomes of these adversarial attacks when applied to the models on the Human 3.6M dataset. The variable $\Delta s$ represents the Euclidean distance between the adversarial examples and their corresponding real examples, and $w/o$ represents the original average MPJPE without any attack.

Fig. \ref{fig:FGSM_epsilon} shows the effect of epsilon on average MPJPE, similar effect occurred in our experiment for IFGSM and MIFGSM. We therefore fixed the  $\epsilon$ value for our experiments in  Tabs. \ref{tab:Attacks_H36M_epsilon_0.001} and \ref{tab:Attacks_H36M_epsilon_0.01} at 0.001 and 0.01. Another reason for that was because of the increase of $\Delta s$ as average MPJPE increases. The goal of the adversarial attacks is to keep $\Delta s$ as small as possible but to maximize average MPJPE. When $\Delta s$ becomes too large, it becomes infeasible to occur in reality. In terms of number of iterations it was specified as 10 for IFGSM, MIFGSM, and DeepFool. The parameter $\mu$ in MIFGSM was set to 0.4. DeepFool has only one parameter, the number of iterations. 

We highlight that previous attempts at adversarial attacks were mostly applied to images, which typically have pixel values normalized to the range between 0 and 1. However, in our case, we are applying adversarial attacks to 3D human pose sequences, which consist of real numbers larger than 1. Therefore, to determine the actual $\epsilon$ value for each sample, we scaled the predefined epsilon value using the expression from Eqs. \ref{eq:fgsm}, \ref{eq:ifgsm}, and \ref{eq:mifgsm2}, while considering the vertical axis, as indicated by the term $| X_{max} - X_{min} |$. This approach ensures that the added perturbation remains within a reasonable and meaningful range.

\begin{figure*}
    \centering
    \subfloat[$\epsilon$ equal to 0.1]{\includegraphics[width=0.33\textwidth, keepaspectratio]{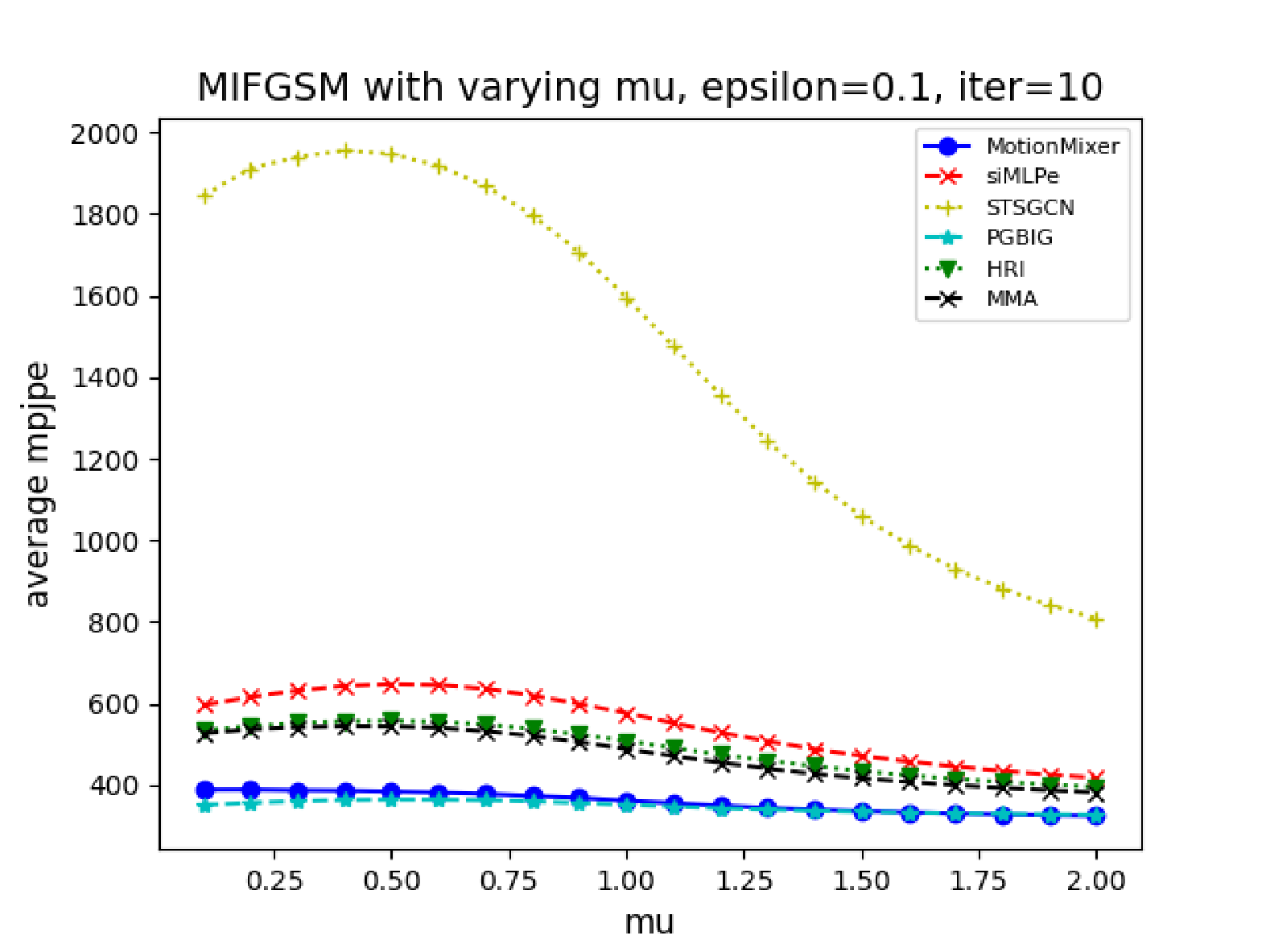}}
    \subfloat[$\epsilon$ equal to 0.01]{\includegraphics[width=0.33\textwidth, keepaspectratio]{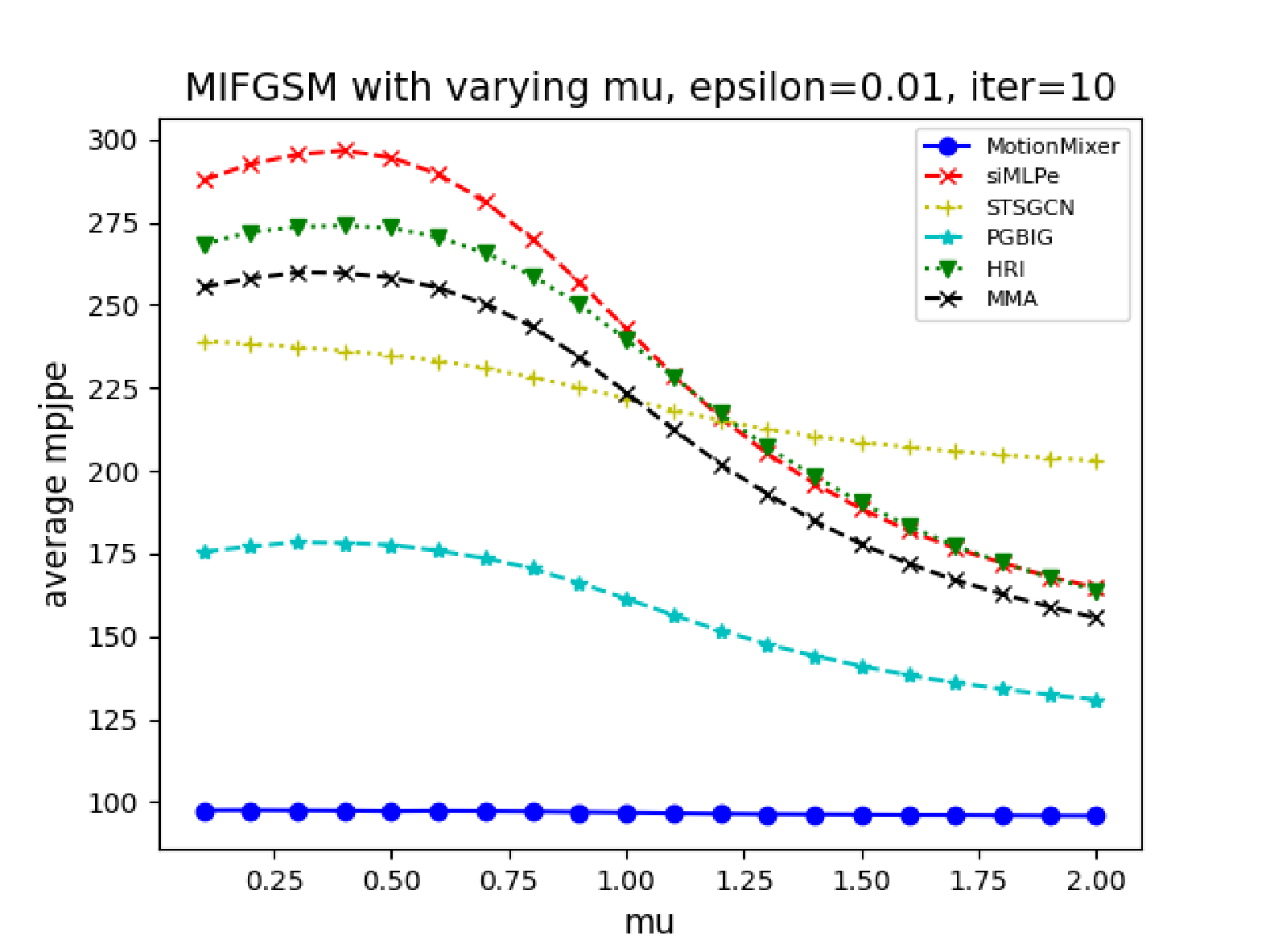}}
    \subfloat[$\epsilon$ equal to 0.001]{\includegraphics[width=0.33\textwidth, keepaspectratio]{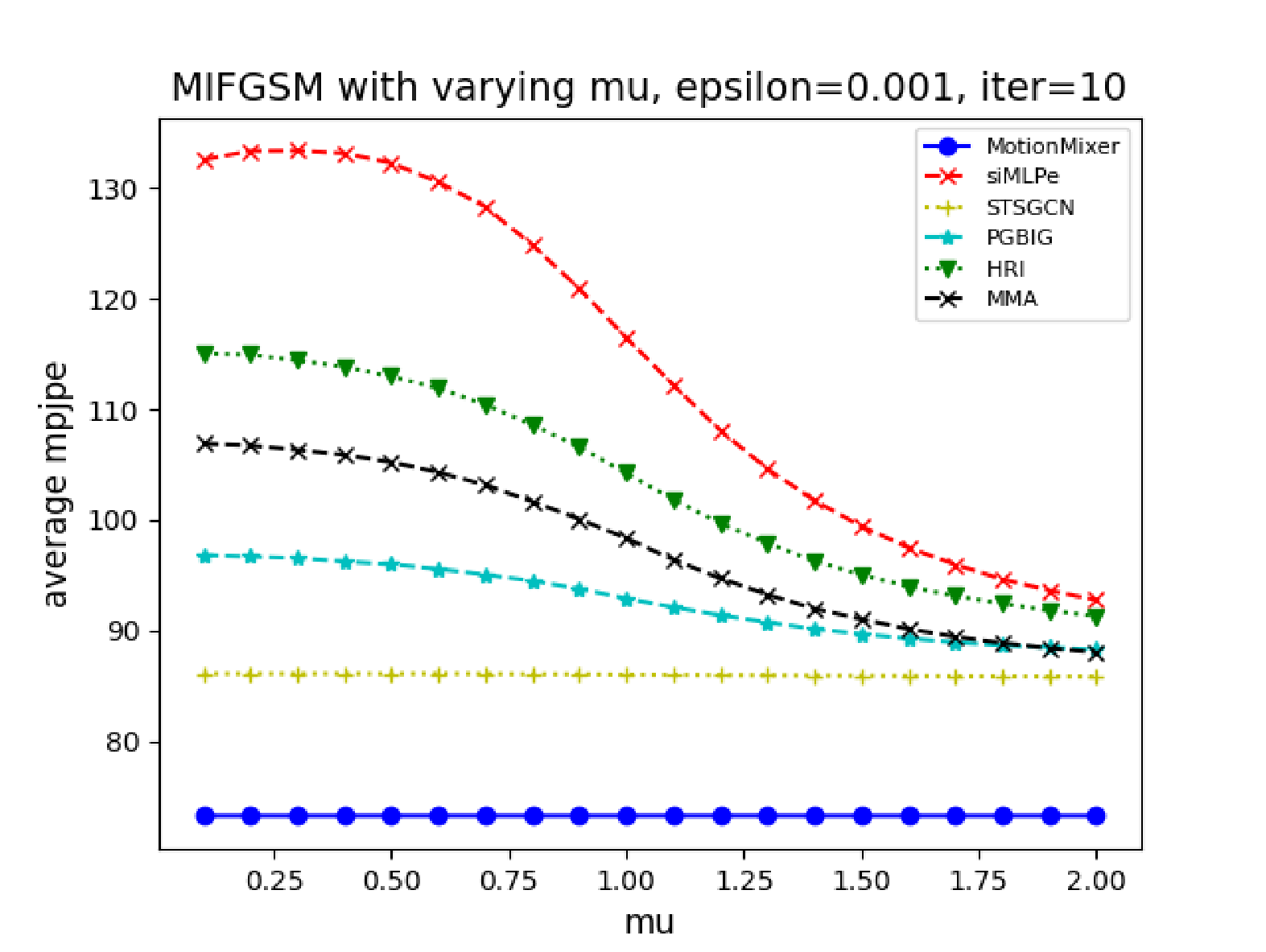}}

    \caption{Result of MIFGSM on average MPJPE with $\mu$ ranging from 0.0 to 2.0 with granularity 0.1 using 10 iterations.}
    \label{fig:MIFGSM_mu}
\end{figure*}

\begin{figure}
	\centering
	\subfloat[Rotations between 0-360 degrees in the Y-axis]{
		\includegraphics[width=0.22\textwidth]{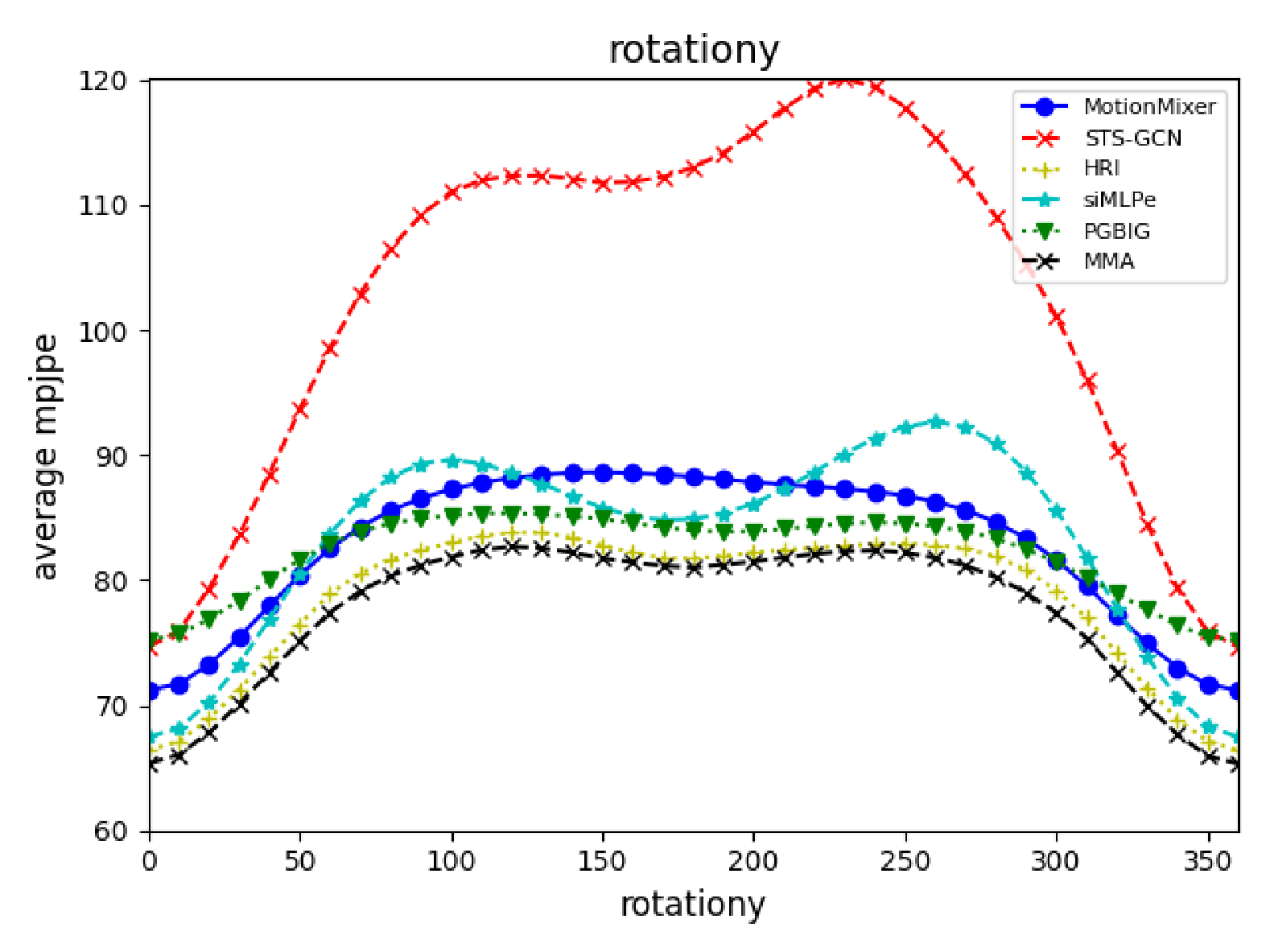}
		\label{fig:rotations}
	}
	\hfill
	\subfloat[Scale rate between 0.7 and 1.3 in the Y-axis]{
		\includegraphics[width=0.22\textwidth]{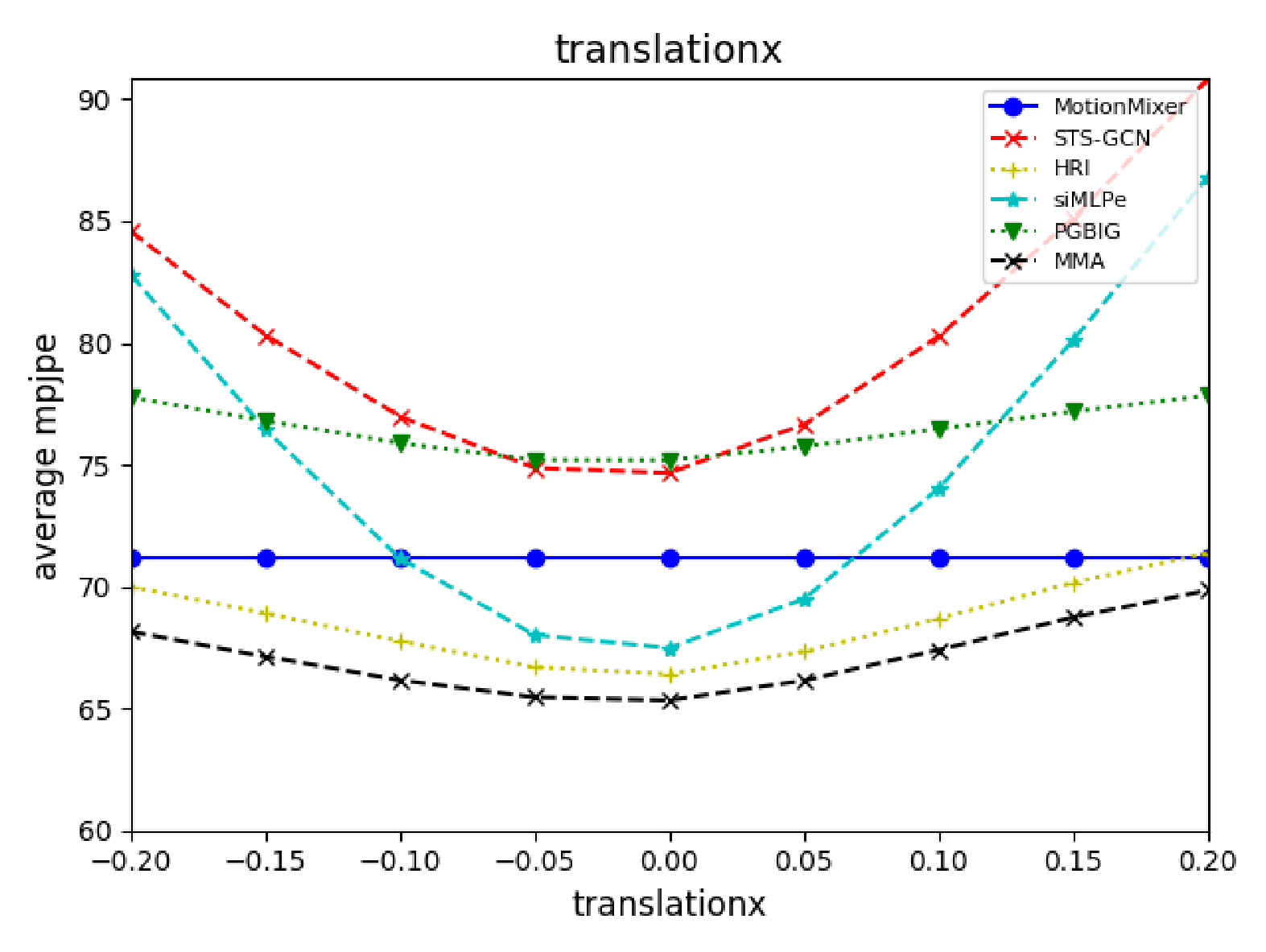}
		\label{fig:scale1}
	}
	\hfill
	\subfloat[Translation rate between -0.2 and 0.2 in the X-axis]{
		\includegraphics[width=0.22\textwidth]{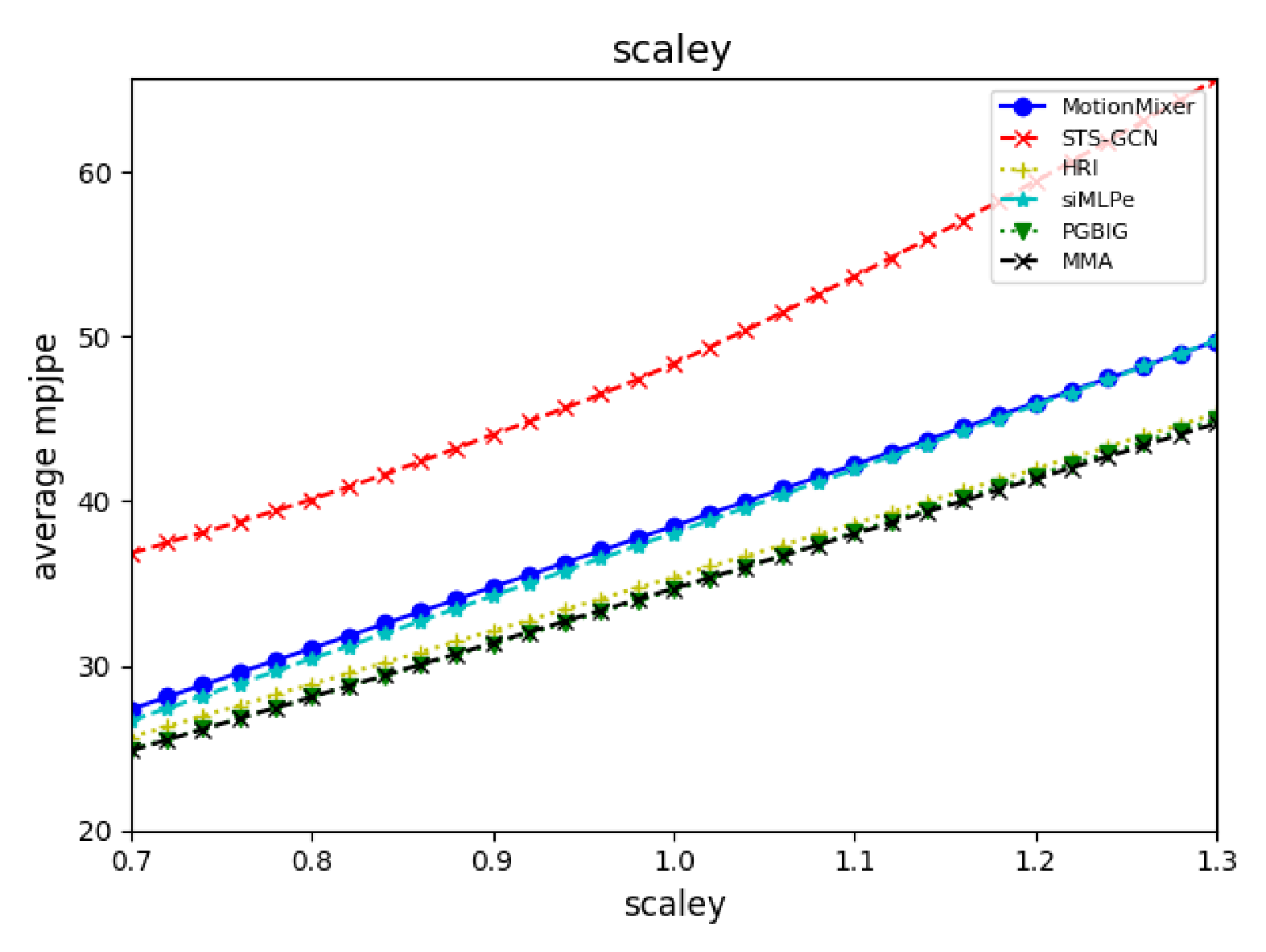}
		\label{fig:translation}
	}
	\hfill
	\subfloat[Scale rate between 0.7 and 1.3 in the X-axis.]{
		\includegraphics[width=0.22\textwidth]{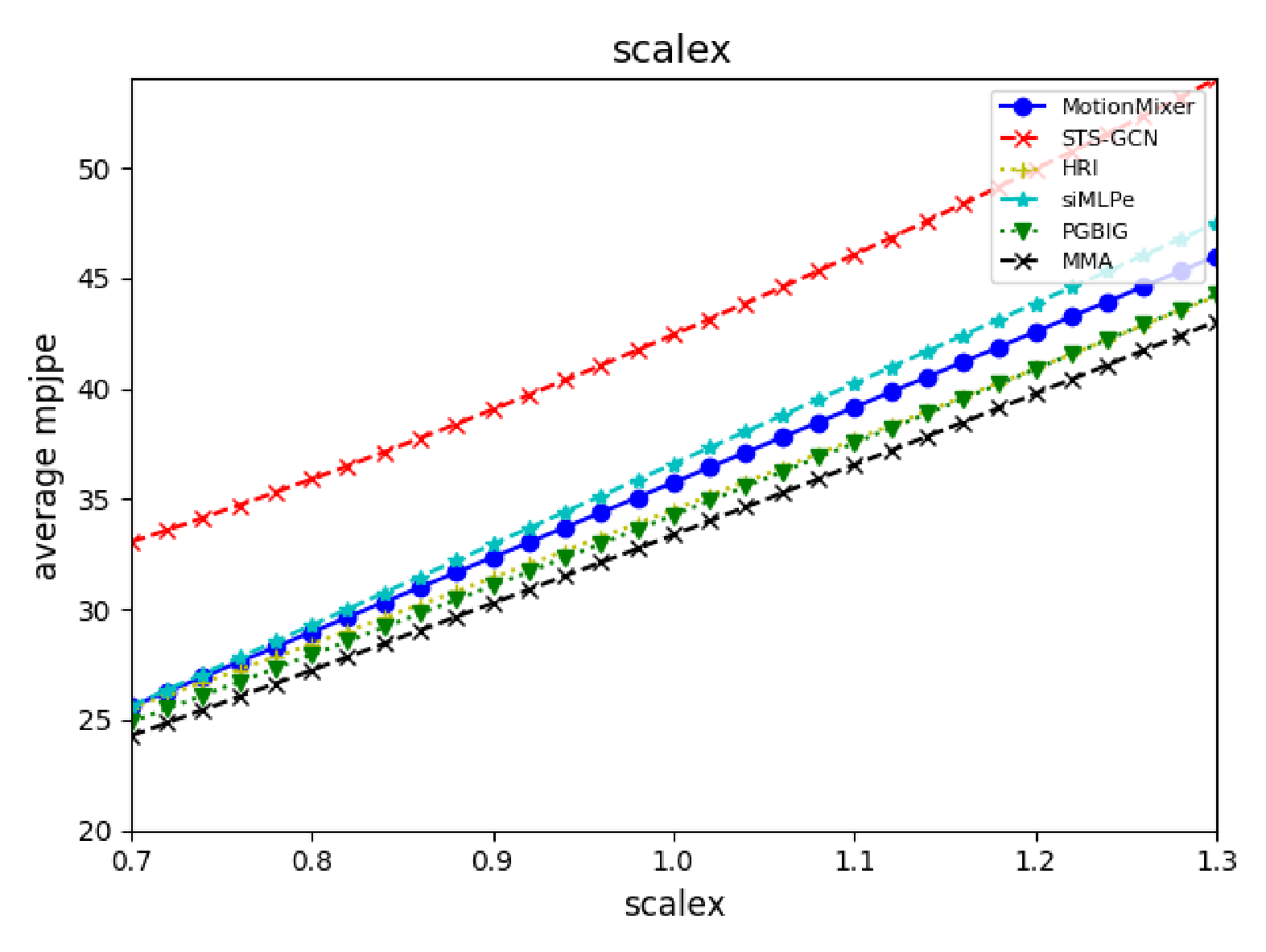}
		\label{fig:scale2}
	}
	\caption{Transformation effects on the test set using the average MPJPE over the 25 output frames}
	\label{fig:SpatialPerturbation}
\end{figure}

\begin{figure}
	\centering
	\subfloat[MotionMixer on 3DPW]{
		\includegraphics[trim=10 0 10 10, clip, width=0.22\textwidth]{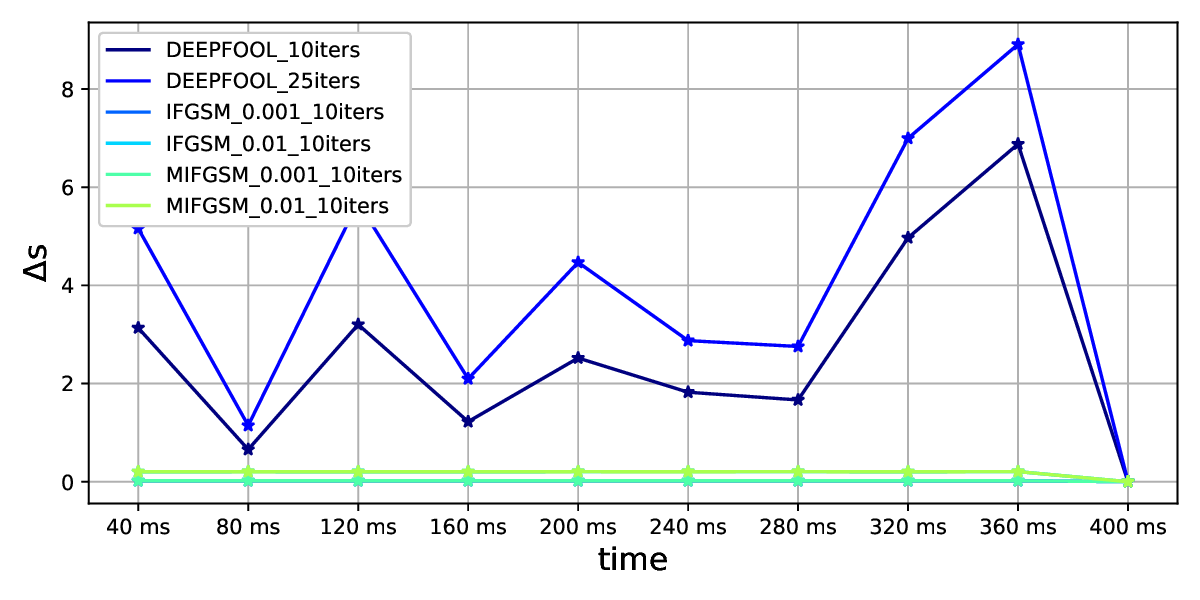}
		\label{fig:motionmixer_diff_3dpw}
	}
	\hfill
	\subfloat[STSGCN on 3DPW]{
		\includegraphics[trim=10 0 10 10, clip, width=0.22\textwidth]{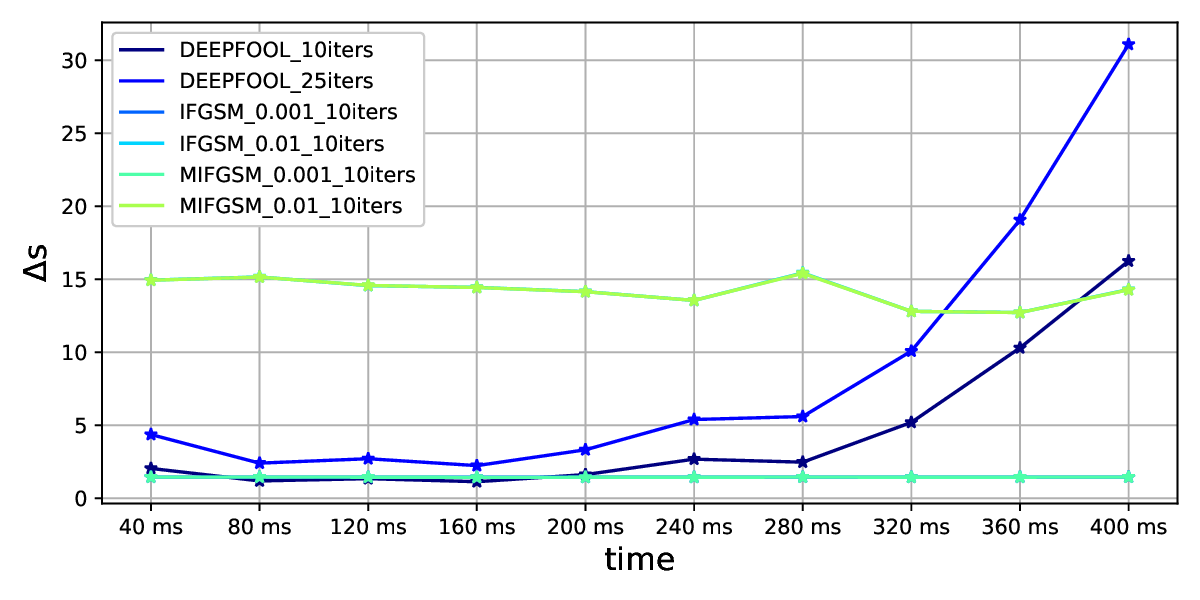}
		\label{fig:stsgcn_diff_3dpw}
	}
	\hfill
	\subfloat[MotionMixer on AMASS]{
		\includegraphics[trim=10 0 10 10, clip, width=0.22\textwidth]{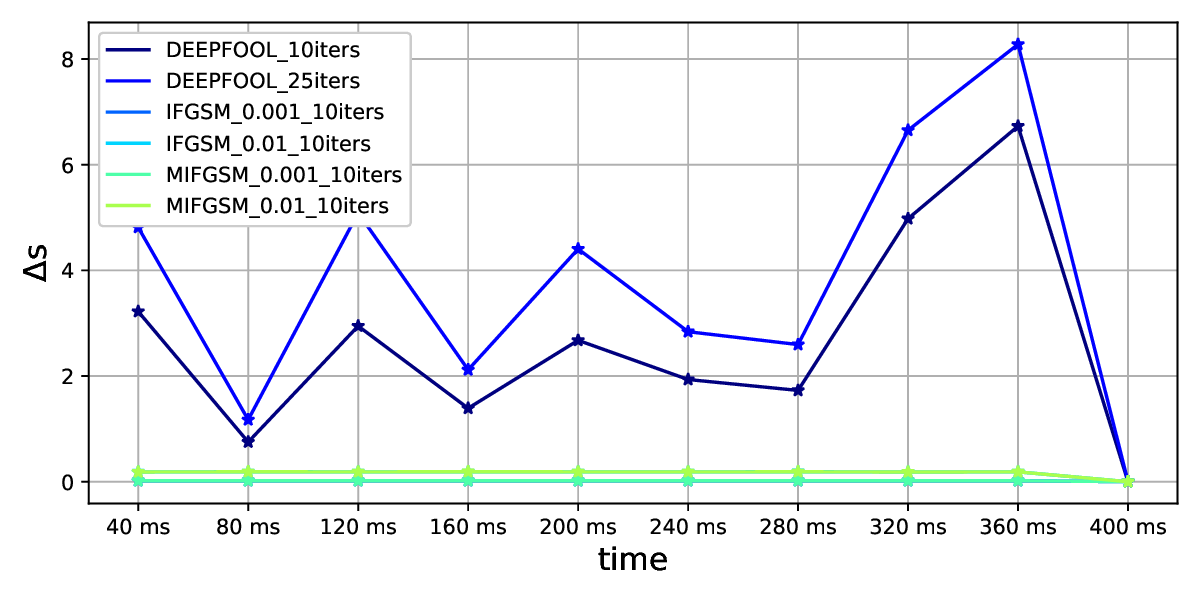}
		\label{fig:motionmixer_diff_amass}
	}
	\hfill
	\subfloat[STSGCN on AMASS]{
		\includegraphics[trim=10 0 10 10, clip, width=0.22\textwidth]{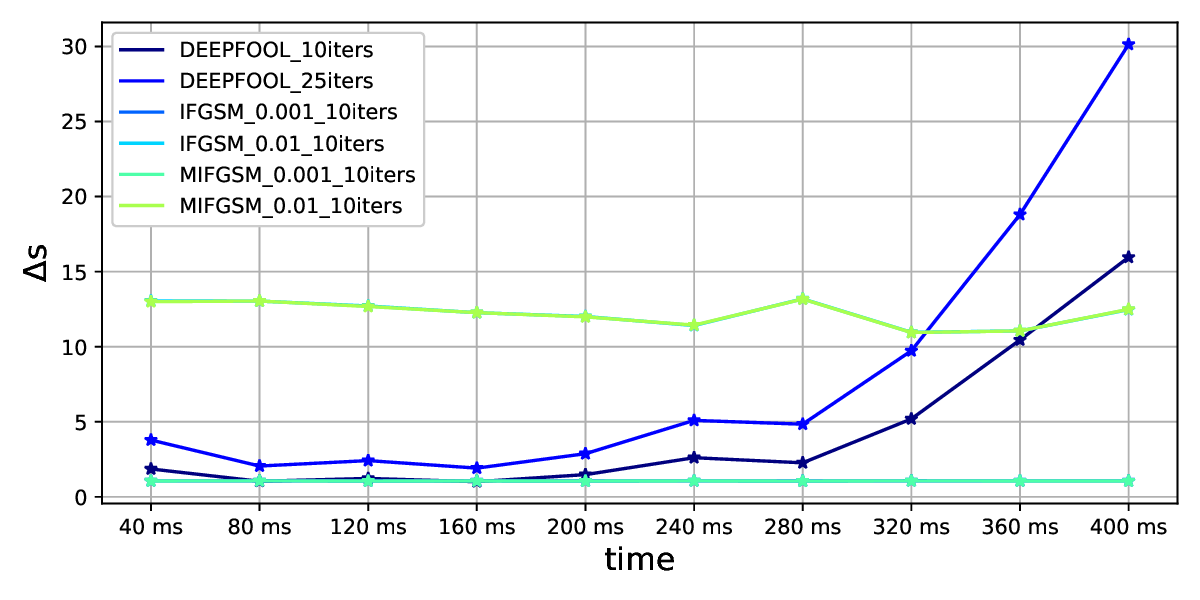}
		\label{fig:stsgcn_diff_amass}
	}
	
	\caption{Adversarial attacks applied to STSGCN and MotionMixer models on the AMASS and 3DPW datasets.}
	\label{fig:deepfool_ablation}
	\vspace{-1mm}
\end{figure}

\begin{figure*}[h]
	\centering
	\includegraphics[width=0.485\textwidth]{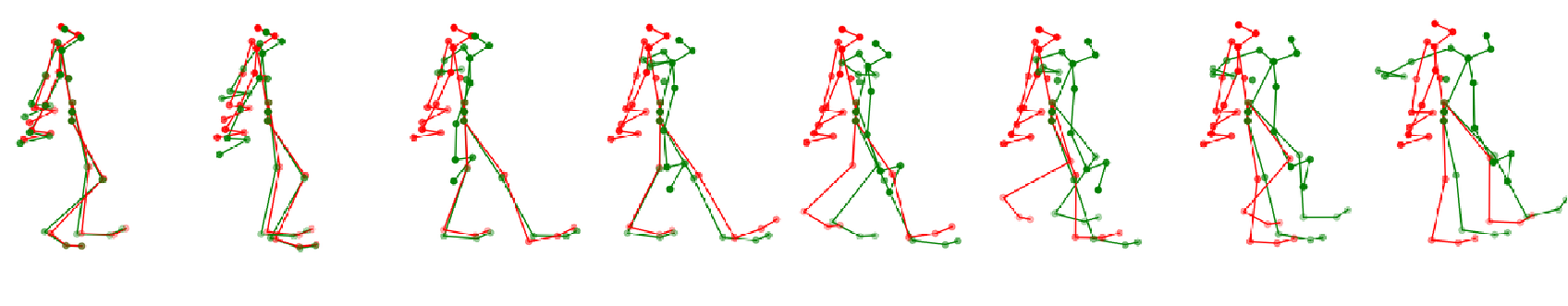}
	\hfill
	\includegraphics[width=0.485\textwidth]{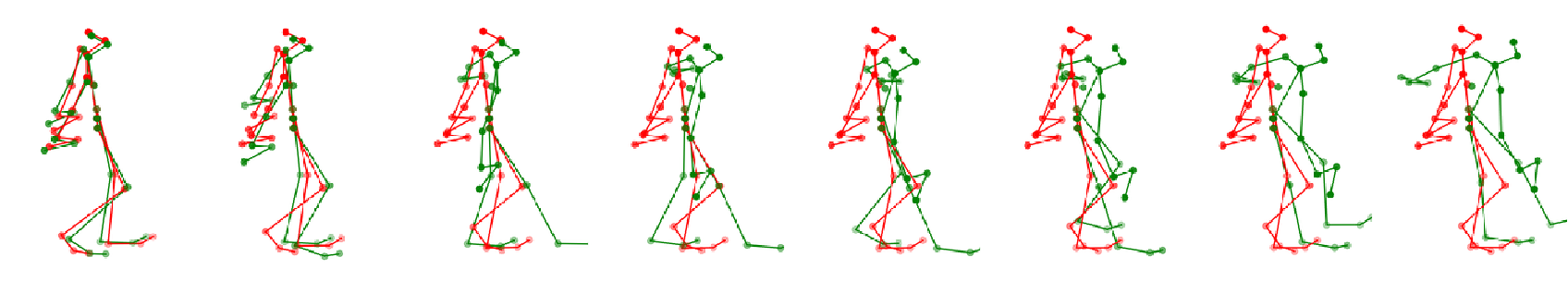}
	\hfill
	\includegraphics[width=0.485\textwidth]{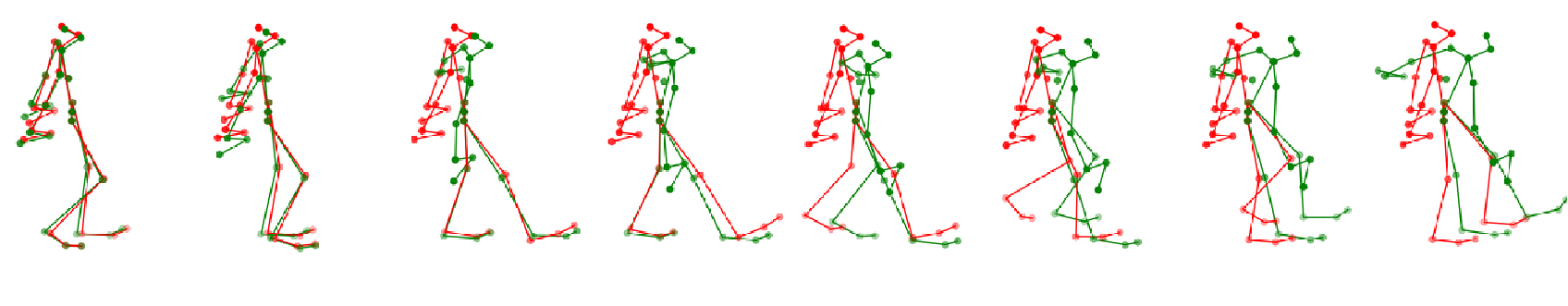}
	\hfill
	\includegraphics[width=0.485\textwidth]{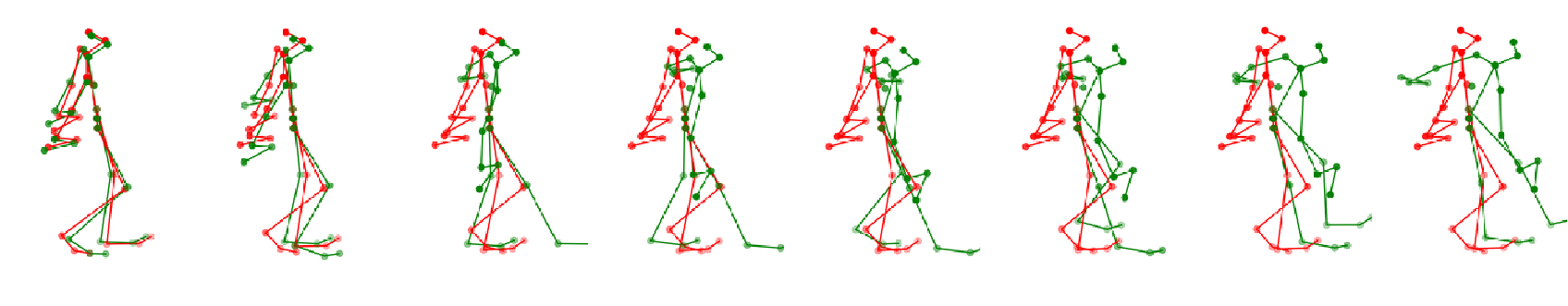}
	\hfill
	\includegraphics[width=0.485\textwidth]{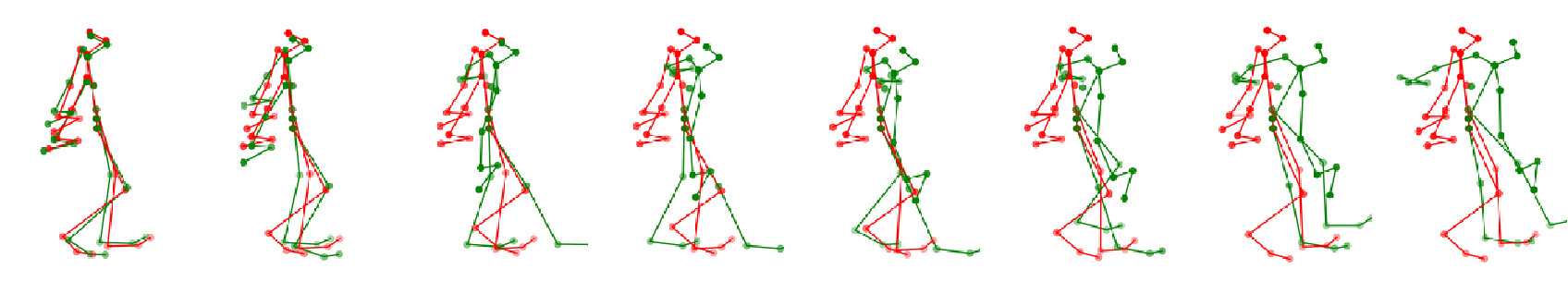}
	\hfill
	\includegraphics[width=0.485\textwidth]{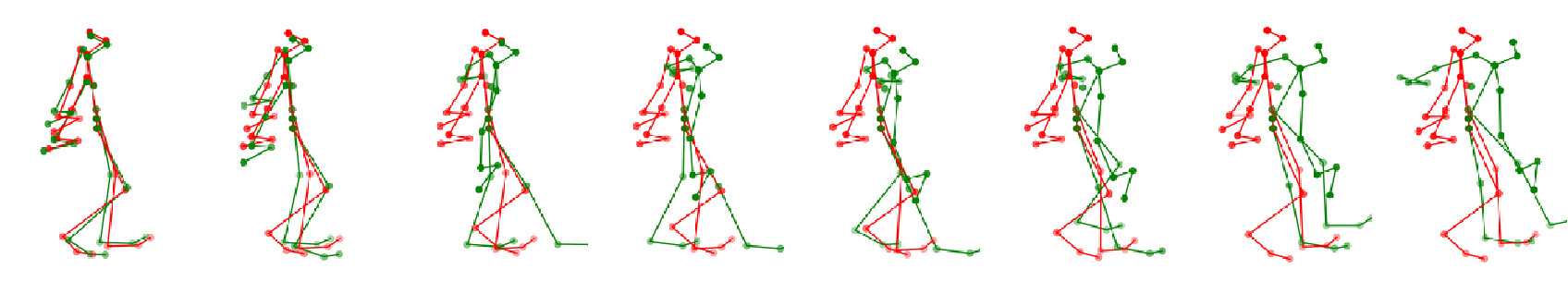}
	\hfill
	\includegraphics[width=0.485\textwidth]{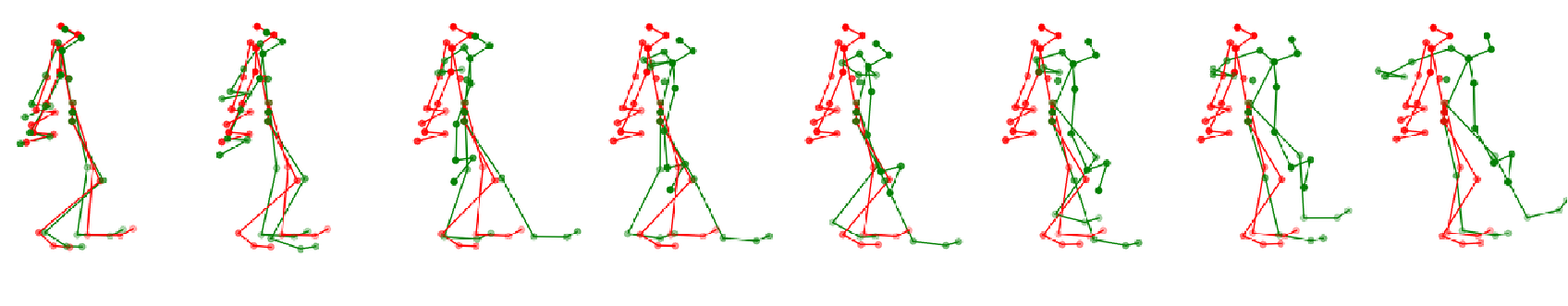}
	\hfill
	\includegraphics[width=0.485\textwidth]{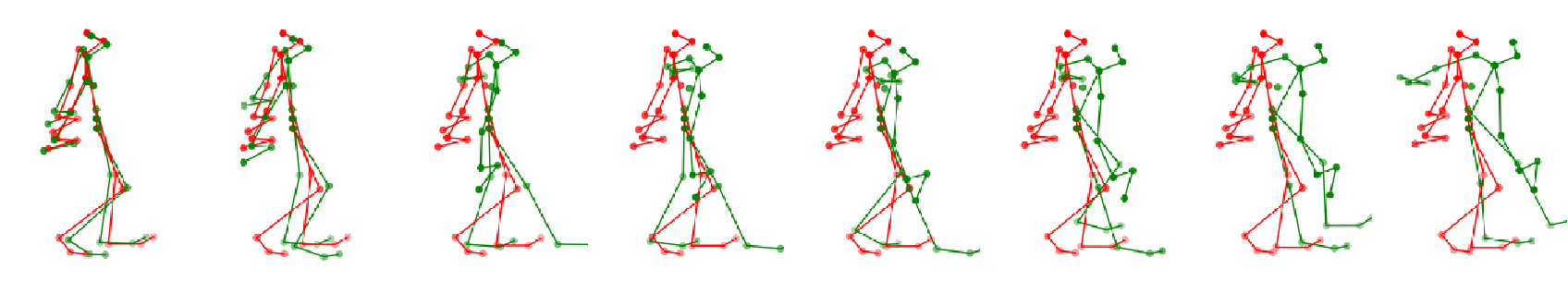}
	\hfill
	\includegraphics[width=0.485\textwidth]{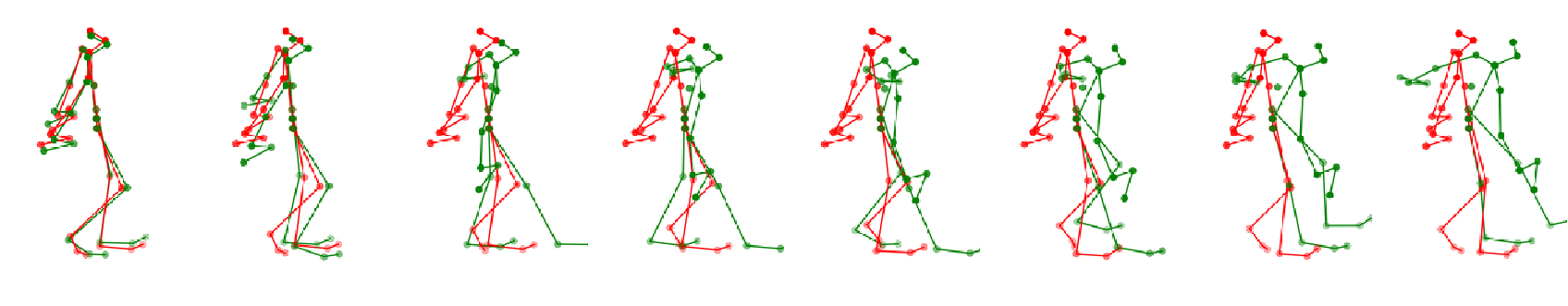}
	\hfill
	\includegraphics[width=0.485\textwidth]{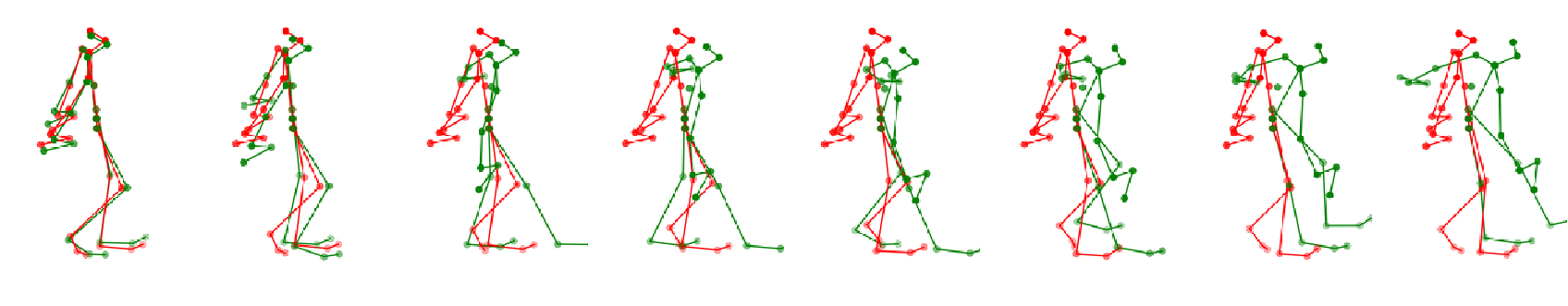}
	\vspace{-2.5mm}
	\hfill
	
	\subfloat[Original input.]{
		\includegraphics[width=0.485\textwidth]{Images/Results/0_STSGCN_walking_000078.eps}
		\label{fig:original_out}
	}
	\hfill
	\subfloat[Rotation 240$^{\circ}$ in the vertical axis.]{
		\includegraphics[width=0.485\textwidth]{Images/Results/240_STSGCN_walking_000078.eps}
		\label{fig:rotated_out}
	}
	
	
	\caption{prediction for a walking pedestrian before and after rotating 240$^{\circ}$.}
	\label{fig:rotation_attack}
\end{figure*}

\begin{figure}
	\centering
	\subfloat[original]{
		\includegraphics[width=0.485\textwidth]{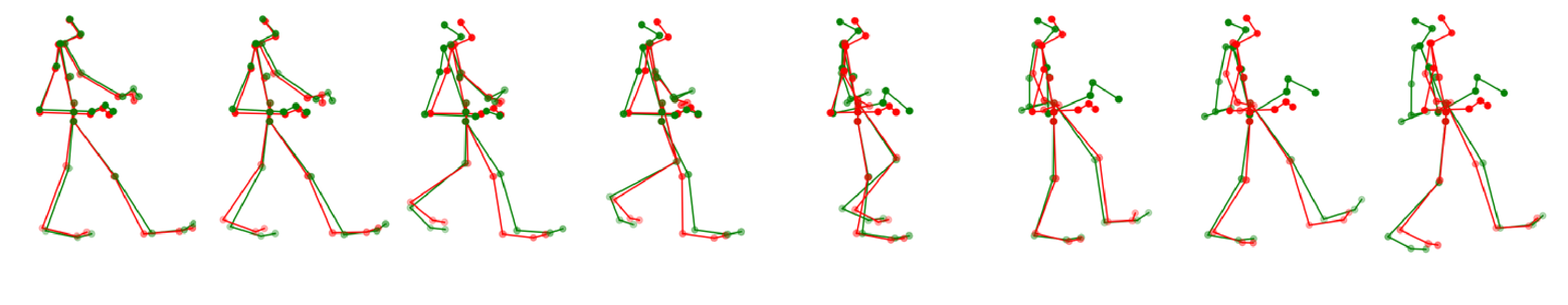}
	}
	\hfill
	\subfloat[$\epsilon=0.001$]{
		\includegraphics[width=0.485\textwidth]{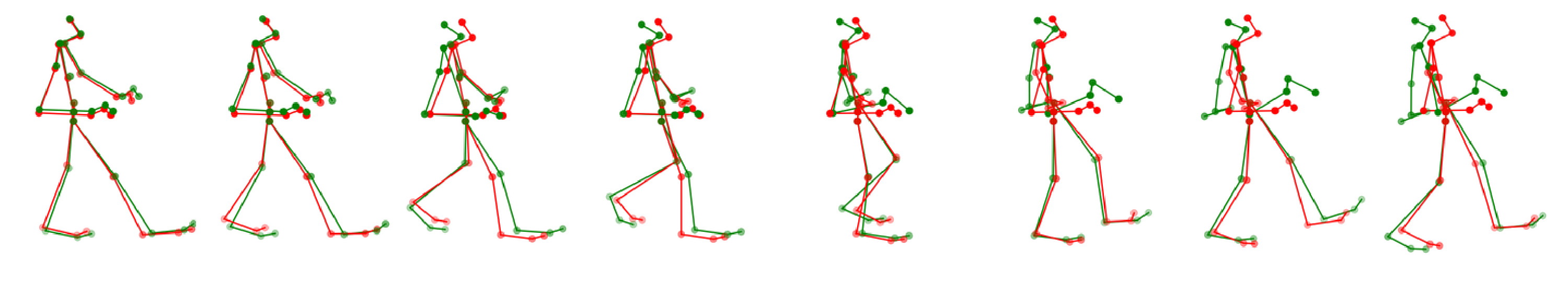}
	}
	\hfill
	\subfloat[$\epsilon=0.01$]{
		\includegraphics[width=0.485\textwidth]{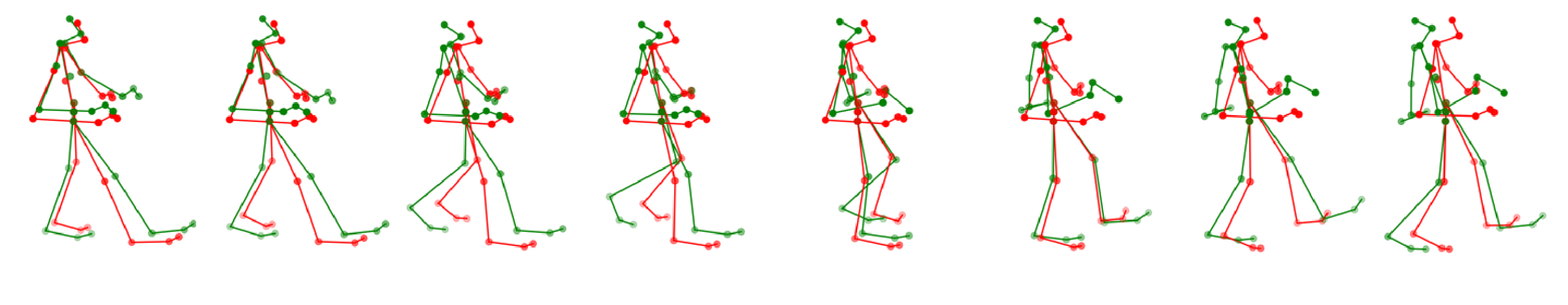}
	}
	\hfill
	\subfloat[$\epsilon=0.1$]{
		\includegraphics[width=0.485\textwidth]{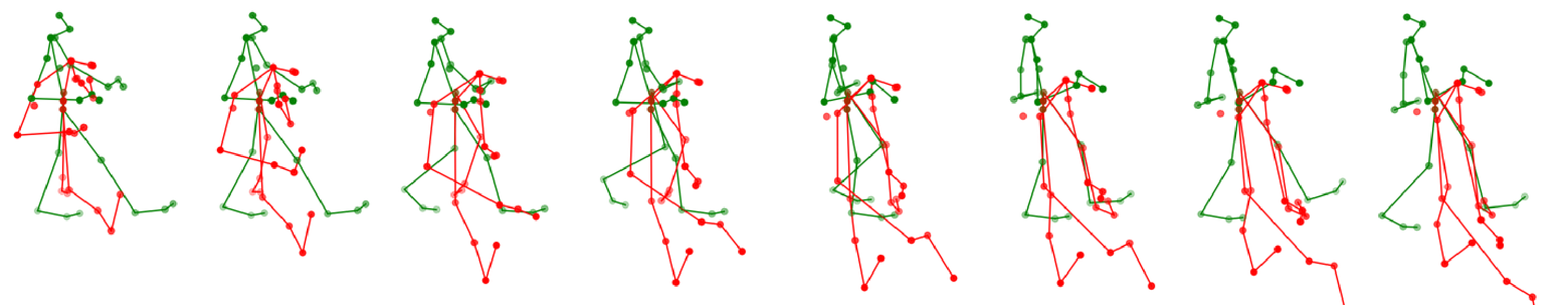}
	}
	\caption{prediction for a walking pedestrian after applying an adversarial attack.}
	\label{fig:ifgsm_attack}
\end{figure}

As shown in Tabs. \ref{tab:Attacks_H36M_epsilon_0.001} and \ref{tab:Attacks_H36M_epsilon_0.01}, MotionMixer stands out as the most robust model in comparison to the others. This heightened robustness can be attributed to the utilization of pose displacements as input data, while other models rely on pose positions or pre-processed versions of these. Consequently, by incorporating changes in pose positions between consecutive frames as input, we increase the robustness of the model. In our ablation study, we explore the impact of the parameter $\mu$ in the MIFGSM method on the MPJPE. Fig. \ref{fig:MIFGSM_mu} reveals that the optimal $\mu$ value falls within the range of 0.25 to 0.5, which consistently yields the largest average MPJPE error across all models. We also noted that the input sequences vary in length among our models, and this variation can also impact the gradient magnitude, adding an additional layer of complexity to our analysis. Also, to demonstrate the performance and generalization capabilities of these attacks. We also apply the same IFGSM, MIFGSM, and DeepFool algorithms to available SOTA models for the AMASS and 3DPW datasets, this is presented in Tabs. \ref{tab:Attacks_AMASS_epsilon_0.01} and \ref{tab:Attacks_3DPW_epsilon_0.01} respectively. As we can observe, MotionMixer reflects the best robustness against attacks. We indicate with ($^\ast$) to point to the models that were trained by us using the author's codes.

In addition to conventional adversarial attacks, we have introduced spatial perturbations to our models and evaluated their performance based on the MPJPE. As depicted in Fig. \ref{fig:SpatialPerturbation}, STS-GCN exhibits sensitivity to spatial perturbation, primarily due to its higher gradient (rate of change concerning perturbation) compared to other models. On the other hand, MotionMixer, which utilizes pose displacements as input, displays a robust behavior to rotation and translation perturbations, as these alterations have minimal impact on the performance following 3D transformations. For this reason, MotionMixer behaves as the most robust architecture against translation perturbation, as reflected by the nearly flat average MPJPE, which remains unchanged for different translation factors applied to the input. Finally, we conducted experiments involving scale transformations to demonstrate that the MPJPE performance does not exhibit a linear increase for all models in a uniform manner. Although we know that the MPJPE metric is influenced by the scale factor, we observe that the model predictions are also not accurately scaled. All the tests presented in this context can be considered as out-of-distribution data.

\textbf{Qualitative results}.
In order to show visually the effect of the adversarial attacks on the models. In Fig. \ref{fig:rotation_attack}, we visualize the output of the SOTA models for a “walking” motion in the following order: HRI, MMA, MotionMixer, PGBIG, STSCGN, and siMLPe. We show the effect of rotation over the vertical axis and how this affects the model performance. More detailed, Fig. \ref{fig:original_out} shows the original prediction while Fig. \ref{fig:rotated_out} shows the prediction for a rotated input. The plot shows the samples with predictions of 80, 160, 320, 400, 560, 720, 880, and 1000 ms. Additionally, we also present in Fig. \ref{fig:ifgsm_attack} the output prediction of STSGCN before (first row) and after applying the IFGSM algorithm for different epsilon values. We show the output prediction at timestamps of 80, 160, 320, 400, 560, 720, 880, and 1000 ms. The average MPJPE for this “walking” sample is originally 48.35 but after applying IFGSM, the values average MPJPE are 53.30, 109.08, and 592.89 for epsilon values at 0.001, 0.01, and 0.1. We know that epsilon at 0.1 is large enough to fool the network but have a cost on the $\Delta s$ in the input domain. Visually we can observed a prediction collapse for $\epsilon$ equal to 0.1 in Fig. \ref{fig:ifgsm_attack}. We also understand that in regression, numerical stability has a large cost when these attack algorithms are applied. But with a defined metric for the difference between the input and adversarial samples $\Delta s$, we could use this controlled noise in the training stage.

\section{\uppercase{Discussion}}
Despite conducting an extensive array of experiments employing SOTA models on well-established datasets, we observed that the prediction error remained nearly consistent across all frames. This phenomenon can be attributed to the threshold imposed by the FGSM algorithm family. However, Deepfool, which utilizes the output predictions and processes them with their gradients, allows for the generation of a non-flat response when introducing adversarial noise. To apply Deepfool effectively in our context with pose sequences, certain adaptations were needed due to disparities in input and output shapes. In contrast to Deepfool, FSGM uses gradients that have the same shape as the input sequence. Consequently, our approach takes the average of the output gradients in the time domain, which was subsequently propagated as a single-frame error. This adaptation resulted in a distinct gradient behavior. This is illustrated in Fig. \ref{fig:deepfool_ablation} for AMASS and 3DPW datasets, where the horizontal axis presents the frame index and the vertical axis presents the difference between the original and adversarial sequences, denoted as $\Delta s$. We use the mean Hausdorff distance metric for MotionMixer and the MPJPE metric for STS-GCN. This choice was made to see more insightful visualizations, particularly considering that MotionMixer employs displacement-based representations with values that are typically very low. It can be observed that the adversarial attacks algorithms learn to exert the most variation in $\Delta s$ to the last frame, in order to fool the models.  Additionally, in the case of MotionMixer, the last displacement is replaced with the positions of the last frame for this evaluation, that's why the $\Delta s$ drop to zero for the last displacement frame.

\section{\uppercase{Conclusions and future direction}}
We observed that models are easily fooled by adversarial attacks as same as in the initial stages of CNNs on image classification. Furthermore, we showed that 3D spatial transformations also behave as no-gradient-based attack methods and have strong effects on the model performance. 
As a future direction, we plan to use these methods as data augmentation for more realistic scenarios such as small short-period rotations or spatio-temporal windowed noise. We also plan to explore black-box methods since we observed white-box attacks worked successfully and also a white-box method using the gradients to guide the 3D spatial transformations.

\section*{\uppercase{Acknowledgements}}
The research leading to these results is funded by the German Federal Ministry for Economic Affairs and Climate Action within the project “ATTENTION – Artificial Intelligence for realtime injury prediction”. The authors would like to thank the consortium for the successful cooperation.

{\small
	\bibliographystyle{apalike}
	\bibliography{foolingnets}{}
}

\end{document}